\documentclass[sigconf]{acmart}
\usepackage{enumitem}
\usepackage{bbm}
\usepackage{multirow}
\usepackage{caption}
\usepackage{graphicx}
\usepackage{float} 
\usepackage{subcaption}
\usepackage{balance}
\usepackage[linesnumbered, ruled]{algorithm2e}
\SetKwRepeat{Do}{do}{while}%

\AtBeginDocument{%
  \providecommand\BibTeX{{%
    \normalfont B\kern-0.5em{\scshape i\kern-0.25em b}\kern-0.8em\TeX}}}

\setcopyright{acmcopyright}
\copyrightyear{2023}
\acmYear{2023}
\setcopyright{acmlicensed}\acmConference[CIKM '23]{Proceedings of the 32nd
ACM International Conference on Information and Knowledge
Management}{October 21--25, 2023}{Birmingham, United Kingdom}
\acmBooktitle{Proceedings of the 32nd ACM International Conference on
Information and Knowledge Management (CIKM '23), October 21--25, 2023,
Birmingham, United Kingdom}
\acmPrice{15.00}
\acmDOI{10.1145/3583780.3614926}
\acmISBN{979-8-4007-0124-5/23/10}

\begin{document}
\newcommand{\our}{\textsc{{iLoRE}}}


\title{{\our}: Dynamic Graph Representation with Instant Long-term Modeling and Re-occurrence Preservation}

\author{Siwei Zhang}
  \email{swzhang22@m.fudan.edu.cn}
  \affiliation{%
    \institution{Shanghai Key Laboratory of Data Science, School of Computer Science, Fudan University}
    \city{Shanghai}
    \country{China}
   }

\author{Yun Xiong}
\authornote{Corresponding author}
 \email{yunx@fudan.edu.cn}
  \affiliation{%
     \institution{Shanghai Key Laboratory of Data Science, School of Computer Science, Fudan University}
     \city{Shanghai}
    \country{China}
  }

\author{Yao Zhang}
  \email{yaozhang@fudan.edu.cn}
  \affiliation{%
     \institution{Shanghai Key Laboratory of Data Science, School of Computer Science, Fudan University}
     \city{Shanghai}
      \country{China}
  }

\author{Xixi Wu}
  \email{21210240043@m.fudan.edu.cn}
  \affiliation{%
    \institution{Shanghai Key Laboratory of Data Science, School of Computer Science, Fudan University}
    \city{Shanghai}
    \country{China}
  }

\author{Yiheng Sun}
  \email{sunyihengcn@gmail.com}
  \affiliation{%
    \institution{Tencent Weixin Group}
    \city{Shenzhen}
    \country{China}
  }

\author{Jiawei Zhang}
  \email{jiawei@ifmlab.org}
  \affiliation{%
     \institution{IFM Lab, Department of Computer Science, University of California, Davis}
     \state{CA}
     \country{USA}
  }

\renewcommand{\shortauthors}{Siwei Zhang, et al.}

\begin{abstract}
Continuous-time dynamic graph modeling is a crucial task for many real-world applications, such as financial risk management and fraud detection. Though existing dynamic graph modeling methods have achieved satisfactory results, they still suffer from three key limitations, hindering their scalability and further applicability.
i) \textbf{Indiscriminate updating.} For incoming edges, existing methods would indiscriminately deal with them, which may lead to more time consumption and unexpected noisy information.
ii) \textbf{Ineffective node-wise long-term modeling.} They heavily rely on recurrent neural networks (RNNs) as a backbone, which has been demonstrated to be incapable of fully capturing node-wise long-term dependencies in event sequences.
iii) \textbf{Neglect of re-occurrence patterns.} Dynamic graphs involve the repeated occurrence of neighbors that indicates their importance, which is disappointedly neglected by existing methods.

In this paper, we present \textbf{{\our}}, a novel dynamic graph modeling method with \textbf{\underline{i}}nstant node-wise \textbf{\underline{Lo}}ng-term modeling and \textbf{\underline{Re}}-occurrence preservation.
To overcome the indiscriminate updating issue, we introduce the Adaptive Short-term Updater module that will automatically discard the useless or noisy edges, ensuring {\our}'s effectiveness and instant ability. We further propose the Long-term Updater to realize more effective node-wise long-term modeling, where we innovatively propose the Identity Attention mechanism to empower a Transformer-based updater, bypassing the limited effectiveness of typical RNN-dominated designs. Finally, the crucial re-occurrence patterns are also encoded into a graph module for informative representation learning, which will further improve the expressiveness of our method. Our experimental results on real-world datasets demonstrate the effectiveness of our {\our} for dynamic graph modeling.
\end{abstract}




\maketitle

\section{Introduction}
\label{sec:intro}
In real-world scenarios, graphs are often constantly evolving over time, where objects (nodes) and their interactions (edges) can emerge and change along a temporal sequence. Such graphs are known as continuous-time dynamic graphs\footnote{For simplicity, we use ``dynamic graph'' in the following text.} \cite{dynamicGraph2022}. Graph Neural Networks (GNNs) for modeling static graphs \cite{GNNs2023, gtransformer_2022, gtransformer2_2022} fail to encode the temporal dependencies, leading to inferior performance when applied to dynamic graphs. Fortunately, Temporal Graph Networks (TGNs) \cite{DyRep2019, JODIE2019, TGAT2020, TGN2020, tiger2023} proposed in recent years effectively learn the temporal representation of dynamic graphs. TGNs focus on developing effective aggregation methods for incorporating historical neighbors, such as self-attention \cite{TGAT2020} and summation \cite{DyRep2019}. Most TGNs utilize a memory module to record nodes' historical behavior, enabling them to make predictions about future events. Despite their effectiveness, existing TGNs still have some key limitations:

\begin{figure}
  \centering
  \includegraphics[width=\linewidth]{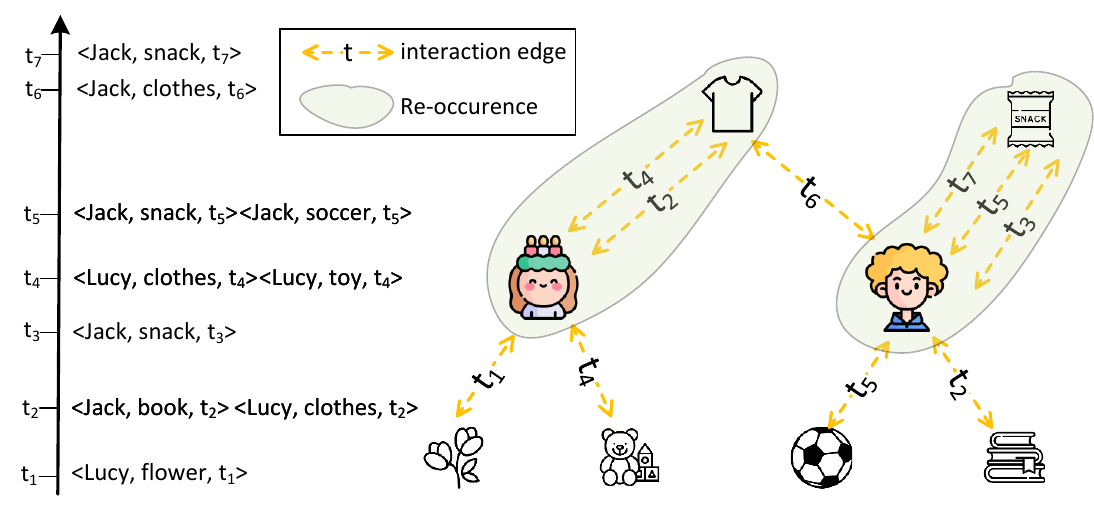}
  \caption{An example of purchase event sequences with time order (left) and the corresponding dynamic graph (right).}
  \label{introduction}
\end{figure}

\textbf{Indiscriminate updating.}
TGNs indiscriminately update the memory of every node by encoding the information from each incoming edge \cite{TGN2020, TGAT2020}. Indiscriminate updating would increase the redundant computational time consumption for the models and also diminish the models' ability to provide results instantly. It significantly limits their deployment in concrete industrial application tasks, especially in those that take instant ability into consideration such as financial risk management or fraud detection \cite{fraud2023, apan2021}. Another issue is that indiscriminate updating may introduce useless or noisy edges \cite{edge2021}, which will further pollute and adversely affect the quality of representation generation. 

\textbf{Ineffective node-wise long-term modeling.}
Unlike the adjacency matrix of static graphs, dynamic graphs are represented as event sequences with time order, as shown in Figure \ref{introduction}. There can be a large number of events occurring around certain nodes, resulting in abundant historical neighbors. These nodes are so-called ``big nodes'' \cite{edge2021}, e.g. Lucy and Jack in Figure \ref{introduction}, and the frequency of updates increases with the number of edges connected to them. Existing methods heavily rely on Recurrent Neural Networks (RNNs) \cite{rnn2020, GRU2014}, and fail to fully capture the node-wise long-term dependencies, particularly in the case of big nodes. Hence, more effective modeling of node-wise long-term dependencies is necessary.

\textbf{Neglect of re-occurrence patterns.}
In dynamic graphs, events around two nodes can occur at different time. As shown in Figure \ref{introduction}, this phenomenon is referred to as ``re-occurrence'', where multiple edges can exist between two nodes. Intuitively, the frequency of re-occurrence can serve as an indication of the importance. For instance, in the purchase dynamic graph, re-occurrence reflexes the interests of consumers. As depicted in Figure \ref{introduction}, Jack, who previously purchased snacks multiple times, is more likely to make another snack purchase in the future. However, TGNs have not leveraged the valuable patterns, limiting their overall effectiveness.

To address the aforementioned limitations, in this paper, we propose a novel dynamic graph modeling method named \textbf{{\our}} (Dynamic Graph Representation with \textbf{\underline{i}}nstant Node-wise \textbf{\underline{Lo}}ng-term Modeling and \textbf{\underline{Re}}-occurrence Preservation). {\our} consists of three main components:
i) Adaptive Short-term Updater.
To estimate the effect of indiscriminate updating, a state module is proposed to adaptively determine the utility of incoming edge for short-term modeling, allowing us to either incorporate or discard it accordingly.
ii) Long-term Updater.
To achieve node-wise long-term modeling, we employ a Transformer-based updater instead of typical RNN-based designs. However, the distribution of a certain node in event sequences is scattered. Applying full attention in such a case will increase the learning difficulty and hampers our ability to capture node-wise long-term dependencies effectively. Therefore, we propose Identity Attention, which can re-sort, pad, chunk, and apply time-aware attention within a chunk in event sequences. For more time-sensitive cases, we employ Gaussian Range Encoding \cite{that2021} and time encoding \cite{TGAT2020} to preserve the temporal information. 
iii) Re-occurrence Graph Module.
To encode the re-occurrence patterns in dynamic graphs, we fetch the re-occurrence number of historical neighbors to indicate their importance to the central node. Specifically, we apply a graph module that leverages crucial re-occurrence features to generate the informative temporal representation for downstream tasks. 

In summary, our main contributions are:
\begin{itemize}[leftmargin=*]
\item We propose a novel dynamic graph modeling method {\our} in this paper. Different from existing TGNs, {\our} focuses on instant node-wise long-term modeling and re-occurrence preservation.
\item We introduce a state module to determine the utility of incoming edges and enable us to selectively discard useless or noisy ones, which ensures instant ability and the effectiveness of our method.
\item We propose Identity Attention, which empowers our Transformer-based updater for node-wise long-term modeling in event sequences. 
\item We incorporate the valuable re-occurrence features with graph module to generate more informative temporal representation.
\item We conduct extensive experiments on dynamic graphs, demonstrating that {\our} has robust performance in various tasks.
\end{itemize}

\section{Related Work}
\subsection{Dynamic Graph Modeling}
As research on dynamic graphs has become increasingly in-depth, dynamic graph modeling has seen rapid development in recent years \cite{cope2021, caw2021, MATA2021, tiger2023}. These methods can be roughly divided into two categories: sequential models and graph models. 

Early works \cite{deepco2016, JODIE2019, DyRep2019} belong to sequential models, which regard dynamic graphs as event sequences, limiting each node to receiving information from at most one-hop historical neighbors. To address this problem, the authors \cite{TGAT2020} present the first graph model that proposes a temporal attention layer to capture information from multi-hop historical neighbors, which achieves perfect results. Subsequently, many graph models emerged such as \cite{TGN2020, tiger2023, apan2021, edge2021, caw2021, pint2022, tgl2022}, further increasing the popularity of dynamic graph modeling. However, recent research has found that the ``graph module'' in graph models is not necessary \cite{tcl2021, graphmixer2023}. In \cite{graphmixer2023}, the authors simply use a Multi-Layer Procedure (MLP) to model one-hop historical neighbors' information and achieve best results than previous graph models, causing researchers to reconsider the necessity of graph modules. 

Currently, there are few models that consider both these two types of methods. Our proposed method is based on Transformer for long-term modeling in event sequences, followed by a graph module with re-occurrence features for representation generation.

\subsection{Transformers for Graph Learning}
Transformer \cite{transformer2017} is an innovative model for processing sequential data. Its self-attention mechanism allows it to perceive longer sequences, which is of great importance in the field of long-sequence modeling. Currently, Transformer has been successfully applied in many fields, such as computer vision \cite{cv1_2020, cv2_2020, cv3_2021}, natural language processing \cite{nlp1_2018, nlp2_2019, nlp3_2018}, and time series prediction \cite{ts1_2019, ts2_2021, ts3_2021}. 

In static graphs, researchers have proposed many Transformer-based methods for static graph modeling \cite{gtransformer_2022, gtransformer1_2022, gtransformer2_2022}. The authors \cite{graph_transformer_layer2020} propose a graph transformer layer with Laplacian Eigenvectors to encode graph structure. In \cite{graph_attention_layer2022}, the authors utilize a graph transformer attention layer to extract information and capture the neighboring correlations, which achieves effective performance.

Currently, most works in the field of dynamic graphs are based on RNNs, and there are few works that use the Transformer as the backbone. Therefore, our proposed model extends Transformer into node-wise long-term modeling in dynamic graphs, opening up new possibilities in the field of dynamic graph modeling.

\section{Notatoin and Terminology Preliminaries}
\begin{definition}\label{def:graph}
    \textbf{Dynamic Graph.}
    A dynamic graph is a graph whose edges contain temporal information, \textit{i.e.}, timestamps. We denote a dynamic graph as a sequence of timestamped evolving graphs $\mathcal{G}=(\mathcal{G}(t_0), \mathcal{G}(t_1), ...)$, where $t_k < t_{k+1}$ and $\mathcal{G}(t_{k+1})$ is generated from $\mathcal{G}(t_k)$ with the edges whose timestamp is $t_{k+1}$. We represent an edge between nodes $i$ and $j$ at time $t$ as a tuple $(i, j, t)$ with an edge feature $\mathbf{e}_{ij}(t)$. 
\end{definition}
A dynamic graph can also be viewed as event sequences $\mathcal{E}$. Each event $(i, j, t_{k+1}) \in \mathcal{E}$ can be seen as the new edge of $\mathcal{G}(t_{k+1})$ compared to $\mathcal{G}(t_{k})$, and all of the events are sorted by timestamps. For the remaining part of this paper, in referring to the incoming dynamic graph edge sequences, we will misuse the terminologies of ``dynamic graph edge set'' and ``event sequence'' interchangeably without distinguishing their differences.

\begin{figure*}
  \centering
  \includegraphics[width=.9\linewidth]{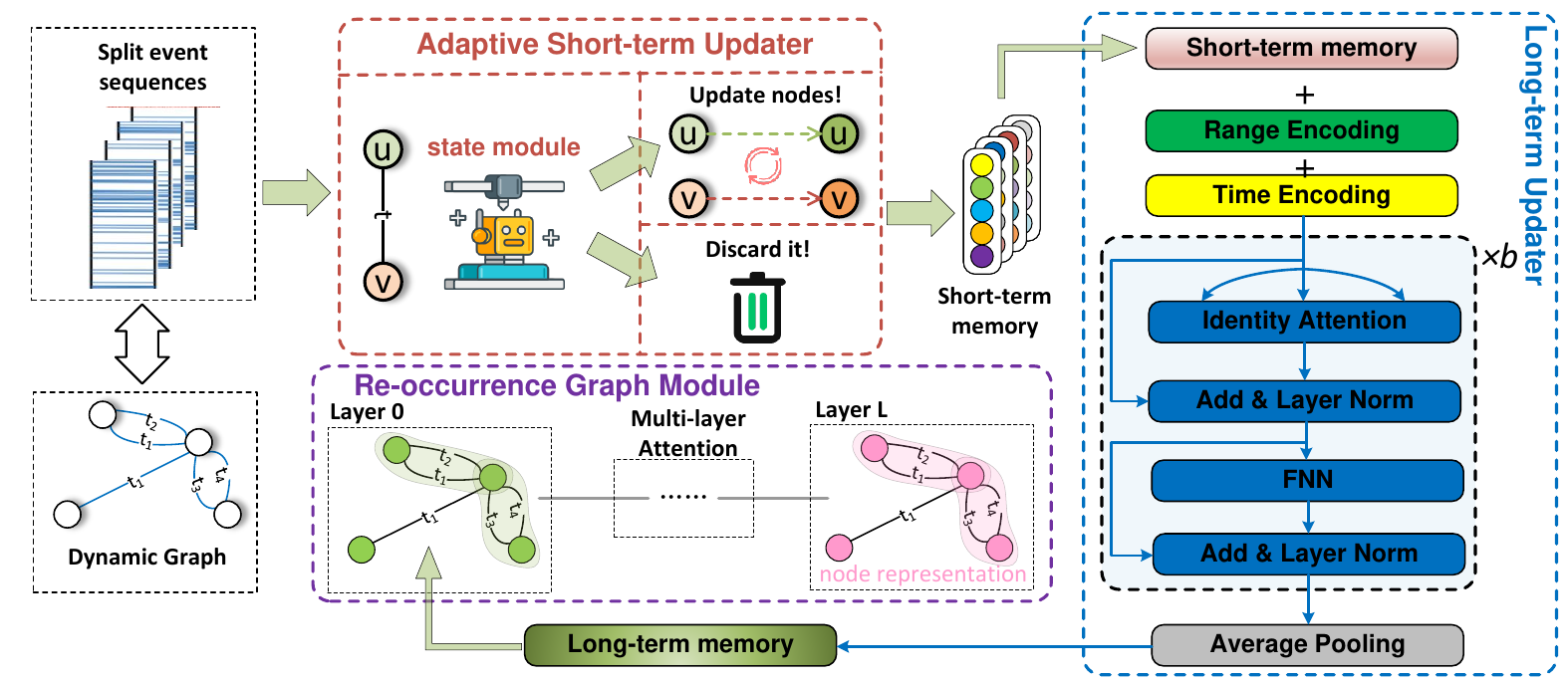}
  \caption{A schematic view of our proposed model for dynamic graph modeling. A dynamic graph can be represented as the event sequences. With the window-split technique in event sequences, in the Adaptive Short-term Updater, a state module is proposed to automatically determine whether to use the incoming event to update nodes for short-term modeling or discard it. Meanwhile, in the Long-term Updater, we propose Identity Attention to empower a Transformer-based updater for node-wise long-term modeling. Gaussian Range Encoding and time encoding are utilized to make our updater more time-sensitive. Finally, we generate the node temporal representation for downstream tasks by applying a multi-layer graph attention module based on the re-occurrence features with nodes' long-term memory.}
  \label{model}
\end{figure*}

\begin{definition}\label{def:modeling}
    \textbf{Dynamic Graph Modeling.}
    Given a dynamic graph edge set or event sequences $\mathcal{E}$, for each event $(i, j, t) \in \mathcal{E}$, the goal of dynamic graph modeling is to learn a mapping function $f: (i, j, t) \mapsto \mathbf{z}_i(t), \mathbf{z}_j(t)$, where $\mathbf{z}_i(t), \mathbf{z}_j(t) \in \mathbbm{R}^d$ respectively represent temporal representation of nodes $i$ and $j$, and $d$ is the vector dimension.
\end{definition}

Besides the terminologies defined above, several other important notations used in this paper are summarized in Table \ref{notations}.
\begin{table}[h]
  \caption{Important notations}
  \label{notations}
  \begin{tabular}{cc}
    \toprule
    Symbol & Definition  \\
    \midrule
    $\mathcal{M}^S_i(t)$ & Short-term memory of node $i$ at $t$ \\
    $\mathcal{M}^L_i(t)$ & Long-term memory of node $i$ at $t$ \\
    $\mathcal{S}_i(t)$ & Node state of node $i$ at $t$ \\
    $\mathbf{X}_{i,\mathcal{R}}(t)$ & Re-occurrence features of node $i$'s neighbors at $t$ \\
    $\mathbf{z}_i(t)$ & Temporal representation of node $i$ at $t$ \\
    \midrule
    $n$ & Chunk size (hyper-parameter) \\
    $b$ & Block number of Transformer (hyper-parameter) \\
  \bottomrule
\end{tabular}
\end{table}

\section{Proposed method}
We first define the short- and long-term behavior of nodes with the window-split technique in event sequences, which are encoded as short- and long-term memory, respectively. Our proposed {\our} has three main parts, including i) the Adaptive Short-term Updater, which achieves instant short-term modeling within a window; ii) the Long-term Updater, which captures nodes' long-term dependencies across multiple windows; iii) and the Re-occurrence Graph Module, which encodes re-occurrence patterns within a graph module for representation. 

As illustrated in Figure~\ref{model}, in the Adaptive Short-term Updater, a state module is proposed to automatically discard useless or noisy edges to ensure the effectiveness and instant ability of our method. Meanwhile, in the Long-term Updater, to empower node-wise long-term modeling ability for event sequences, Identity Attention is proposed to optimize the Transformer-based updater, which can re-sort, pad, chunk, and apply time-aware attention within a chunk. For more time-sensitive cases, we employ Gaussian Range Encoding~\cite{that2021} and time encoding \cite{TGAT2020} to preserve the temporal information. What's more, in the Re-occurrence Graph Module, we incorporate the valuable re-occurrence features into a graph attention module for informative temporal representation generation. We will introduce these components in the following subsections.

\subsection{Node-wise Short- and Long-term Modeling}
\subsubsection{Window-split Technique.}\label{sec:short-long}
To perform node-wise long-term modeling, we propose to split the event sequence into subsequences according to a pre-defined window size $s$. Given event sequences $\mathcal{E} = \{e_1, e_2, ..., e_r\}$ where $r$ is the event length, we define a window set $\mathbf{w} = \{w_1, w_2, ..., w_{\lceil r / s \rceil}\}$, where $w_i = \{e_{i \cdot s - s +1}, e_{i \cdot s - s + 2}, ..., e_{i \cdot s} | \\ i  \le \lceil r / s \rceil\}$ contains $s$ events. 

In this paper, we use short- and long-term memory, $\mathcal{M}^S$ and $\mathcal{M}^L$, to embed the short- and long-term behavior of each node, respectively. The memory of each node $i$, $\mathcal{M}^S_i$ and $\mathcal{M}^L_i$, is initialized as the zero vector and will be updated over time. 
For given node $i$ at time $t$, we use the events within the same window to perform short-term modeling, \textit{i.e.}, updating $\mathcal{M}^S_{i}(t)$, and perform long-term modeling across multiple windows, \textit{i.e.}, updating $\mathcal{M}^L_{i}(t)$. Note that once the long-term memory is updated, the short-term memory will be reset to zero.

\subsubsection{Message Generation.}\label{sec:message} Given an event of node $i$ at time $t$ in window $w_i$, a message $\mathbf{m}_i(t)$ is generated to update the short-term memory of $i$, $\mathcal{M}^S_i(t)$. Assume that nodes $i$ and $j$ have an event at time $t$, $(i, j, t)$, with the feature vector $\mathbf{e}_{ij}(t)$, we generate two messages with the long-term memory of $i$ and $j$, $\mathcal{M}^L_i(t)$ and $\mathcal{M}^L_j(t)$:
\begin{equation}\label{message}
\begin{split}
    \mathbf{m}_i(t) = \textsc{Msg}\left(\mathcal{M}^L_i\left(t\right), \mathcal{M}^L_j\left(t\right), \mathbf{e}_{ij}\left(t\right), \Phi\left(t - t_i^-\right)\right), \\
    \mathbf{m}_j(t) = \textsc{Msg}\left(\mathcal{M}^L_j\left(t\right), \mathcal{M}^L_i\left(t\right), \mathbf{e}_{ij}\left(t\right), \Phi\left(t - t_j^-\right)\right),    
\end{split}
\end{equation}
where $\textsc{Msg}(\cdot)$ is the message function and $t_{*}^-$ is the time that node $i/j$ last updated. $\Phi(\cdot)$ is the time encoding used in \cite{TGN2020}. The reason why we conduct long-term memory for message generation is that it contains more expressive and valuable information compared with short-term one. For simplicity, we implement the widely-used \emph{identity} message function that outputs the inputs. Moreover, in each window $w_i$, we apply the simplest \emph{most recent} message aggregator that only considers the most recent message for each node~\cite{TGN2020}.

\begin{figure*}
  \centering
  \includegraphics[width=.95\linewidth]{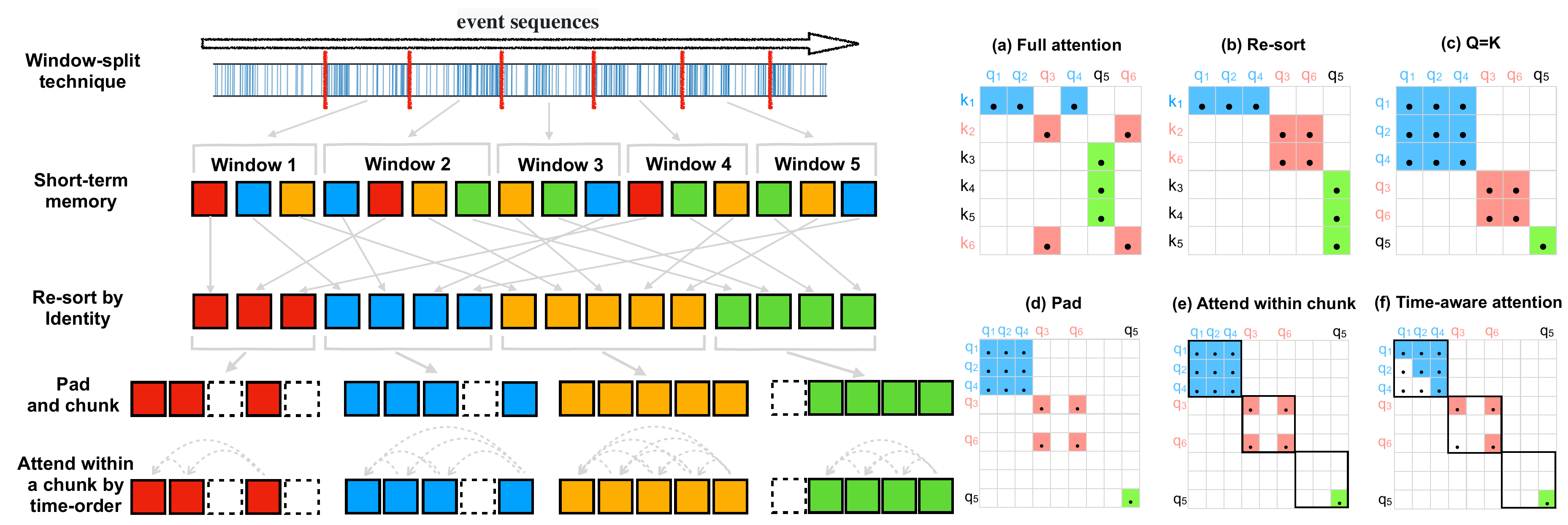}
  \caption{Simplified description of Identity Attention (left) and attention matrices that need to be learned in each step (a-f on right). With the window-split technique, we can take node-wise long-term modeling in multiple widows, e.g., 5 windows, with nodes' short-term memory. Note that different colors denote different node identities. The Identity Attention re-sorts, pads, chunks, and attends within a chunk with time order, which can densify the attention matrix in a chunk, greatly reducing the difficulty to learn the attention matrix and thus increasing our ability to capture node-wise long-term dependencies effectively.}
  \label{identity-attention}
\end{figure*}

\subsection{Adaptive Short-term Updater}
\label{sec:short-term}
To model the node-wise short-term behavior meanwhile ensuring instant ability, we \emph{adaptively update} the short-term memory of node $i$ before $t$, $\mathcal{M}^S_i(t^-)$, with the message of $i$ at $t$, $\mathbf{m}_i(t)$. We propose the node state module.

In this module, each node $i$ has a node state at time $t$, $\hat{\mathcal{S}}_i(t) \in (0, 1)$, which is evolving along with timestamps. We have:
\begin{equation}\label{bernoulli}
    \mathcal{S}_i(t) = \operatorname{Bernoulli}\left(\hat{\mathcal{S}}_i\left(t\right)\right),
\end{equation}
where $\operatorname{Bernoulli}(\cdot)$ denotes sampling from a Bernoulli distribution parameterized by $\hat{\mathcal{S}}_i(t)$, and $\mathcal{S}_i(t) \in \{0, 1\}$.

Then, $\mathcal{S}_i(t)$ is utilized to determine whether we update the short-term memory of node $i$ before time $t$, $\mathcal{M}^S_i(t^-)$:
\begin{equation}\label{short-update}
    \mathcal{M}^S_i(t) = \mathcal{S}_i(t) \cdot \textsc{Upd}\left(\mathcal{M}^S_i\left(t^-\right), \mathbf{m}_i\left(t\right)\right) + \left(1 - \mathcal{S}_i\left(t\right)\right) \cdot \mathcal{M}^S_i\left(t^-\right)
\end{equation}
where $\textsc{Upd}(\cdot)$ is a learnable update module for node-wise short-term modeling, and we use GRU \cite{GRU2014} in practice. Afterward, we update the node state with its short-term memory in the following timestamps, $t^+$:
\begin{align} \label{state-update}
    & \Delta\hat{\mathcal{S}}_i(t) = \sigma \left(\mathbf{W}_p \cdot \mathcal{M}^S_i\left(t\right) + \mathbf{b}_p\right) \\
    & \hat{\mathcal{S}}_i(t^+) = \left(1 - \mathcal{S}_i\left(t\right)\right) \cdot \Delta\hat{\mathcal{S}}_i(t) + 
    \mathcal{S}_i(t) \cdot \left( \hat{\mathcal{S}}_i\left(t\right) - \alpha \operatorname{min}\left(\Delta \hat{\mathcal{S}}_i(t), \mathcal{S}_i(t)\right)\right) 
\end{align}
where $\mathbf{W}_p$ and $\mathbf{b}_p$ are learnable parameters, $\sigma(\cdot)$ is the sigmoid function, and $\alpha \in (0, 1)$ is a control hyper-parameter that ensures the node state is positive. The node state module encodes the observation that the likelihood of a new update operation decreases with the frequency of node-wise updating. Whenever $\mathcal{M}^S_i(t)$ updates, the pre-activation of the node state for the following timestamp, $\hat{\mathcal{S}}_i(t^+)$, is decreased by $\Delta\hat{\mathcal{S}}_i(t)$. On the other hand, if the update is omitted, the accumulated value is flushed and $\hat{\mathcal{S}}_i(t^+) = \Delta\hat{\mathcal{S}}_i(t)$. In this way, we can selectively update the nodes with incoming edges, ensuring effectiveness and instant ability.

\subsection{Long-term Updater}
We consider both node-wise short- and long-term modeling in this paper. As mentioned in section \ref{sec:short-term}, the recent behavior of node $i$ at time $t$ in window $w_i$ is recorded by short-term memory, $\mathcal{M}^S_i(t)$, with the window-split technique. In this section, we introduce a Transformer-based updater that can embed the node-wise long-term behavior, \textit{i.e.}, updating $\mathcal{M}^L$, using the short-term memory in multiple windows.

\subsubsection{Gaussian Range Encoding and time encoding.}
\label{sec:Gaussian}
The order is pretty important in event sequences. Most Transformer-based methods use positional encoding \cite{transformer2017} that is defined on a single point: They employ a highly discriminative encoding for every single point. It can not align with the nature of time in event sequences because the timestamps are continuous. To make the model more order-aware, we use a \emph{range-based} encoding method. Therefore, we employ Gaussian Range Encoding \cite{that2021}. 

Formally, we propose $\mathbf{B} \in \mathbbm{R}^{d \times k }$ as the normalized weights from $k$ Gaussian distributions, where $d$ denotes the dimension of the input vector. It can be shown as follows:
\begin{equation}
    \label{position-encoding}
    \mathbf{B} = \operatorname{softmax}\left(B\right),
\end{equation}
where $B \in \mathbbm{R}^{d \times k}$ is a matrix whose attributes are sampled from $k$ ranges. In matrix $B$, each cell $b_{ij}$ shows the contribution of the $j$-th Gaussian ranges for position $i$, which can be represented as:
\begin{equation}
    \label{position1}
    b_{i j}=-\frac{\left(i-\mu^{(j)}\right)^{2}}{2 \sigma^{(j) 2}}-\log \left(\sigma^{(j)}\right),
\end{equation}
where $\mu^{(j)}$ and $\sigma^{(j)}$ are the mean and standard deviation of $j$-th Gaussian ranges, respectively. For implementation, we set these two parameters to be learnable. Then, the Gaussian Range Embedding is generated by adding the range embeddings to the input vector $X$:
\begin{equation}
    \label{position2}
    \textsc{Gaussian}(X) = X + \mathbf{B} \cdot \mathbf{E},
\end{equation}
where $\mathbf{E} \in \mathbbm{R}^{k \times d}$ is a learnable matrix. This approach uses $k$ learnable Gaussian ranges to express different positions, which makes our position encoding more continuous. Moreover, we adopt classic time encoding \cite{TGAT2020} widely used in dynamic graph modeling to better preserve temporal information.

\subsubsection{Identity Attention}\label{sec:identity-attention}
Figure \ref{identity-attention} illustrates the motivation and the process of Identity Attention. Figure \ref{identity-attention}a expresses the attention matrix that needs to be learned when using full attention for node-wise long-term modeling, where different colors of $k$ and $q$ represent different nodes’ identities. Since the distribution of a node at different times is scattered in the sequence, the attention matrix for full attention is typically \emph{sparse}, making it difficult to learn. Therefore, we propose Identity Attention, which can \emph{densify} the attention matrix within a chunk by re-sorting (Figure \ref{identity-attention}b, c), padding (Figure \ref{identity-attention}d), chunking (Figure \ref{identity-attention}e), and attending within a chunk (Figure \ref{identity-attention}f), greatly reducing the learning difficulty and enhancing our ability to node-wise long-term modeling in event sequences.

We first rewrite the equation of full attention. For a query position $i$, its attention to position $j$ can be represented as $\mathbf{o}_{ij}$:
\begin{equation}\label{attention1}
    \mathbf{o}_{ij}= \sum_{j \in \mathcal{P}_{i}} \exp \left(q_{i} \cdot k_{j}+\mathbf{z}\left(i, \mathcal{P}_{i}\right)\right) v_{j},
\end{equation}
where $\mathcal{P}_{i} = \{j : j \le i \; or \; j > i \}$. Note that $\mathcal{P}_i$ represents the set that the query position $i$ can attend to, and $\mathbf{z}$ denotes the partition function, e.g., $\operatorname{softmax}$. Notably, we omit the parameter $\sqrt{d_k}$ \cite{transformer2017}.

We can use a positional encoding function $\mathbf{m}(\cdot, \cdot)$ to fit the attention between position $i$ and position $j$ that $i$ can attend to:
\begin{equation}\label{attention2}
    \mathbf{o}_{ij}= \sum_{j \in \mathcal{P}_{i}} \exp \left(q_{i} \cdot k_{j}+\mathbf{m}\left(j, \mathcal{P}_i\right)+\mathbf{z}\left(i, \mathcal{P}_{i}\right)\right) v_{j},
\end{equation}
where $\mathbf{m}(\cdot, \cdot)$ usually applies a single point positional encoding~\cite{transformer2017}.

Now we turn to Identity Attention, which we can consider as the constraint of $\mathcal{P}_i$.

\textbf{Re-sort.} 
This step aims to cluster temporal nodes in the same identity. We use bucket sort to rearrange the entire sequence according to the order of identity, where position $i$ changes after sorting, \textit{i.e.}, $i \mapsto c_i$. In the sorted attention matrix, nodes with the same identity will be clustered, as shown in Figure \ref{identity-attention}b. We have:
\begin{equation}\label{attention3}
    \mathcal{P}_{i} = \left\{j : \mathbf{Id}\left(q_i\right) = \mathbf{Id}\left(k_j\right)  \right\},
\end{equation}
where $\mathbf{Id}(\cdot)$ denotes the identity of a correlated temporal node. For simplicity, we let $Q=K$ as represented in Figure \ref{identity-attention}c.

\textbf{Pad and chunk.} 
Since the frequency of node updating is different in different windows, the number of temporal nodes in each bucket is unequal. In practice, we employ zero vectors to pad the temporal nodes that do not have updated in corresponding windows as depicted in Figure \ref{identity-attention}d. Moreover, we apply batching approaches to chunk and concentrate attention within each chunk (after sorting and padding), shown in Figure \ref{identity-attention}e. Formally, we have:
\begin{equation}\label{attention4}
    \widetilde{\mathcal{P}}_{i}=\left\{j:\left\lfloor\frac{c_{i}}{n}\right\rfloor-1 \leq\left\lfloor\frac{c_{j}}{n}\right\rfloor \leq\left\lfloor\frac{c_{i}}{n}\right\rfloor\right\},
\end{equation}
where $n\in \mathbbm{R}^+$ is a hyper-parameter that represents the chunk size. In Long-term Updater, $n$ is also the number of windows where Transformer performs long-term modeling at once in event sequences.

\textbf{Attend within a chunk by time-order.} Then, we implement Gaussian Range Encoding and time-aware attention within a chunk as illustrated in Figure \ref{identity-attention}f. Formally, we also use $\mathbf{m}(\cdot, \cdot)$ as our positional encoding function, which is defined as:
\begin{equation}
\label{attention5}
\mathbf{m}\left(j, \widetilde{\mathcal{P}}_{i}\right)=\left\{
\begin{aligned}
&\textsc{Gaussian}\left(i, \widetilde{\mathcal{P}}_{i}\right), &\operatorname{if} j \in \widetilde{\mathcal{P}}_{i} \\
&-\infty, &\operatorname{otherwise},
\end{aligned}
\right.
\end{equation}
where $\textsc{Gaussian}(\cdot, \cdot)$ is Gaussian Range Encoding in Section \ref{sec:Gaussian}. Considering that in event sequences, the event that happened at time $t$ can only attend to the past events before $t$, we propose time-aware attention, which is defined as:
\begin{equation}
\label{attention6}
\mathbf{t}\left(j, \widetilde{\mathcal{P}}_{i}\right)=\left\{
\begin{aligned}
&\textsc{Time}(i, t), &\operatorname{if} j \in \widetilde{\mathcal{P}}_{i} \operatorname{and} j \le i \\
&-\infty, &\operatorname{otherwise},
\end{aligned}
\right.
\end{equation}
where $\textsc{Time}(\cdot, \cdot)$ is the time encoding in Section \ref{sec:Gaussian}.

Summarily, the final Identity Attention can be represented as:
\begin{equation}
    \label{attention7}
    \mathbf{o}_{ij}= \sum_{j \in \widetilde{\mathcal{P}}_{i}} \exp \left(q_{i} \cdot k_{j}+\mathbf{m}\left(j, \widetilde{\mathcal{P}}_i\right)+\mathbf{t}\left(j, \widetilde{\mathcal{P}}_{i}\right)+\mathbf{z}\left(i, \widetilde{\mathcal{P}}_{i}\right)\right) v_{j}.
\end{equation}
Each component that is negative infinity will force our attention to zero. Similar to full attention, we can also apply the multi-head technique in Identity Attention.

\subsubsection{Transformer}\label{sec:transformer} We employ a standard Transformer encoder for node-wise long-term modeling. Transformer is equipped with stacking $b$ Multi-head Identity Attention (MIA) and Feed-Forward Network (FFN) blocks. Each block employs a residual connection. We use ReLU between the two MLPs in each FFN block and apply Layer Normalization (LN) before each block. 

Thanks to the window-split and chunk technique, the input of Transformer is the short-term memory $\mathcal{M}^S$ that is updated by the events in recent $n$ windows before $w_i$, \textit{i.e.}, $\left\{w_{i-n+1}, ..., w_{i}| i \le \lceil r / s \rceil \right\}$, and it is denoted as $\mathbf{Z}^{0} \in \mathbbm{R}^{l_i \times d}$ where $l_i$ is the length of events. The output embedding of the $b$-th layer is denoted by $\boldsymbol{H} = \boldsymbol{Z}^{b} \in \mathbbm{R}^{l'_i\times d}$ where $l'_i$ is the length of sequence after padding. 

The long-term memory of node $i$ at time $t$, $\mathcal{M}^L_{i}(t)$, is derived by averaging their related embedding in $\boldsymbol{H}$:
\begin{equation}\label{long-term-memory}
    \mathcal{M}^L_i(t) = \operatorname{MEAN}\left(\boldsymbol{H}\left[i, :\right]\right) \in \mathbbm{R}^{d}.
\end{equation}

\begin{algorithm}[t]
  \SetKwInOut{Input}{input}\SetKwInOut{Output}{output}
  \Input{Dynamic graph edge set $\mathcal{E}$; Short-term memory $\mathcal{M}^S$; Long-term memory $\mathcal{M}^L$; Node state $\hat{\mathcal{S}}$; Chunk size $n$.}
  Initialize $\mathcal{M}^S, \mathcal{M}^L, \hat{\mathcal{S}} \leftarrow \mathbf{0}$ \; 
  \ForEach{batch $\{(i, j, t)\} \subseteq \mathcal{E}$}{
    Split batch into $n$ windows, $\{w_1, ..., w_n\}$ \;
    Initialize $\mathcal{M}(t) \leftarrow \mathbf{0}$ \;
    \ForEach{$w_i \in \{w_1, ..., w_n\}$}{
        Sample $\mathcal{S}(t) \sim \operatorname{Bernoulli}(\hat{\mathcal{S}}(t))$ \;
        Update short-term memory $\mathcal{M}^S(t)$ by Equation \ref{short-update} \;
        Update node state $\hat{\mathcal{S}}(t)$ by Equation \ref{state-update} \;
        Record $\mathcal{M}^S(t)$ to $\mathcal{M}(t)$ \;
    }
    Update $\mathcal{M}^L(t) \leftarrow \textsc{Upd}(\mathcal{M}(t))$ with Identity Attention \;
    Compute $\mathbf{X}_{i, \mathcal{R}}(t), \mathbf{X}_{j, \mathcal{R}}(t)$ by Equation \ref{re-occur} \;
    Compute $\mathbf{z}_i(t), \mathbf{z}_j(t) \leftarrow \textsc{Emb}\left(\mathcal{M}^L(t), \mathbf{X}_{i, \mathcal{R}}, \mathbf{X}_{j, \mathcal{R}}\right)$ \;
    Compute $p_{ij}(t), p_{ik}(t)$ by Equation \ref{loss1} \;
    Compute temporal link prediction loss $\mathcal{L}$ by Equation \ref{loss2} and backward \;
  }
  \caption{Traning {\our} (one epoch).}
  \label{algorithm}
\end{algorithm}

\begin{table*}[t]
    \caption{\centering Average Precision $(\text{AP}(\%) \pm \text{Std})$ for temporal link prediction in transductive and inductive setting. The result \textbf{\%d} that is bolded is the best result and the second is \underline{\%d}.} 
    \label{linkprediction}
    \begin{subtable}{.5\linewidth}
      \centering
        \caption{Transductive Setting.}
        \setlength{\tabcolsep}{1.2mm}{
        \begin{tabular}{ccccc}
            \toprule
            & \textbf{Wikipedia} & \textbf{Reddit} & \textbf{MOOC} & \textbf{LastFM} \\
            \midrule
             \textbf{CTDNE}   & 79.42 $\pm$ 0.4 & 73.76 $\pm$ 0.5 & 65.34 $\pm$ 0.7 & 57.25 $\pm$ 1.0 \\
             \textbf{JODIE}   & 94.62 $\pm$ 0.5 & 91.11 $\pm$ 0.3 & 76.50 $\pm$ 1.8 & 68.77 $\pm$ 3.0 \\
             \textbf{TGAT}    & 95.34 $\pm$ 0.1 & 98.12 $\pm$ 0.2 & 60.97 $\pm$ 0.3 & 53.36 $\pm$ 0.1 \\
             \textbf{DyRep}   & 94.59 $\pm$ 0.2 & 97.98 $\pm$ 0.1 & 75.37 $\pm$ 1.7 & 68.77 $\pm$ 2.1 \\
             \textbf{TGN}     & 98.46 $\pm$ 0.1 & 98.70 $\pm$ 0.1 & 85.88 $\pm$ 3.0 & 80.69 $\pm$ 0.2 \\
             \textbf{CAW}     & 98.63 $\pm$ 0.1 & 98.39 $\pm$ 0.1 & 80.15 $\pm$ 0.3 & 81.29 $\pm$ 0.1 \\
             \textbf{TIGER}   & \underline{98.38 $\pm$ 0.1} & \underline{99.04 $\pm$ 0.1} & \underline{89.64 $\pm$ 0.9} & 87.85 $\pm$ 0.9\\
             \textbf{GraMixer}& 97.95 $\pm$ .03 & 97.31 $\pm$ .01 & 82.78 $\pm$ 0.2 & 67.27 $\pm$ 2.1 \\
             \textbf{PINT}    & 98.78 $\pm$ 0.1 & 99.03 $\pm$ .01 & 85.14 $\pm$ 1.2 & \underline{88.06 $\pm$ 0.7} \\
             \textbf{Ours}& \textbf{98.98 $\pm$ 0.3} & \textbf{99.11 $\pm$ 0.4} & \textbf{90.44 $\pm$ 1.0} & \textbf{91.39 $\pm$ 0.1} \\
            \bottomrule
        \end{tabular}}
    \end{subtable}%
    \begin{subtable}{.5\linewidth}
      \centering
        \caption{Inductive Setting.}
        \setlength{\tabcolsep}{1.2mm}{
        \begin{tabular}{ccccc}
            \toprule
            & \textbf{Wikipedia} & \textbf{Reddit} & \textbf{MOOC} & \textbf{LastFM} \\
            \midrule
             \textbf{CTDNE}   & - & - & - & - \\
             \textbf{JODIE}   & 93.11 $\pm$ 0.4 & 94.36 $\pm$ 1.1 & 77.83 $\pm$ 2.1 & 82.55 $\pm$ 1.9 \\
             \textbf{TGAT}    & 93.99 $\pm$ 0.3 & 96.62 $\pm$ 0.3 & 63.50 $\pm$ 0.7 & 55.65 $\pm$ 0.2 \\
             \textbf{DyRep}   & 92.05 $\pm$ 0.3 & 95.68 $\pm$ 0.2 & 78.55 $\pm$ 1.1 & 81.33 $\pm$ 2.1 \\
             \textbf{TGN}     & 97.81 $\pm$ 0.1 & 97.55 $\pm$ 0.1 & 85.55 $\pm$ 2.9 & 84.66 $\pm$ 0.1 \\
             \textbf{CAW}     & 98.24 $\pm$ .03 & 97.81 $\pm$ 0.1 & 81.42 $\pm$ 0.2 & 85.67 $\pm$ 0.5 \\
             \textbf{TIGER}   & \underline{98.45 $\pm$ 0.1} & \underline{98.39 $\pm$ 0.1} & \underline{89.51 $\pm$ 0.7} & 90.14 $\pm$ 1.0 \\
             \textbf{GraMixer}& 96.65 $\pm$ .02 & 95.26 $\pm$ .02 & 81.41 $\pm$ 0.2 & 82.11 $\pm$ 0.4 \\
             \textbf{PINT}    & 98.38 $\pm$ .04 & 98.25 $\pm$ .04 & 85.39 $\pm$ 1.0 & \underline{91.76 $\pm$ 0.7} \\
             \textbf{Ours}& \textbf{98.60 $\pm$ 0.3} & \textbf{98.65 $\pm$ 0.3} & \textbf{89.75 $\pm$ 0.8} & \textbf{93.29 $\pm$ 0.8} \\
            \bottomrule
        \end{tabular}}
    \end{subtable} 
\end{table*}

\begin{table*}[t]
  \caption{\centering AUC $(\text{AUC}(\%) \pm \text{Std})$ for evolving node classification task on Wikipedia, Reddit and MOOC. The result \textbf{\%d} that is bolded is the best result and the second is \underline{\%d}.}
  \label{evolvingClassification}
  \begin{tabular}{cccccccccc}
    \toprule
     & \textbf{CTDNE} & \textbf{JODIE} & \textbf{TGAT} & \textbf{DyRep} & \textbf{TGN} & \textbf{TIGER} & \textbf{GraMixer} & \textbf{PINT} & \textbf{Ours} \\
    \midrule
    \textbf{Wikipedia} & 75.89 $\pm$ 0.5 & 84.84 $\pm$ 1.2 & 83.69 $\pm$ 0.7 & 84.59 $\pm$ 2.2 & \underline{87.81 $\pm$ 0.3} & 86.92 $\pm$ 0.7 & 86.80 $\pm$ .01 & 87.59 $\pm$ 0.6 & \textbf{91.37 $\pm$ 0.2}  \\
    \textbf{Reddit}    & 59.43 $\pm$ 0.6 & 61.83 $\pm$ 2.7 & 65.56 $\pm$ 0.7 & 62.91 $\pm$ 2.4 & 67.06 $\pm$ 0.9 & \underline{69.41 $\pm$ 1.3} & 64.22 $\pm$ .03 & 67.31 $\pm$ 0.2 & \textbf{71.82 $\pm$ 1.6}  \\
    \textbf{MOOC}      & 67.54 $\pm$ 0.7 & 66.87 $\pm$ 0.4 & 53.95 $\pm$ 0.2 & 67.76 $\pm$ 0.5 & 69.54 $\pm$ 1.0 & \underline{72.35 $\pm$ 2.3} & 67.21 $\pm$ .02 & 68.77 $\pm$ 1.1 & \textbf{73.89 $\pm$ 2.0} \\
    \bottomrule
  \end{tabular}
\end{table*}

\subsection{Re-occurrence Graph Module}\label{sec:re-occurrence}
We aim to encode the re-occurrence features into the graph module, which refers to the property that two nodes may interact at different timestamps. Intuitively, the re-occurrence number of a historical neighbor indicates its importance to the central node. 

For given a node $i$ and its historical neighbors at time $t$, $\mathcal{N}_i\left(t\right)$, we count the number of re-occurrence of each neighbor, which is represented as $\mathcal{R}_i\left(t\right) \in \mathbb{R}^{|\mathcal{N}_i\left(t\right)|\times 1}$. Then, we apply a function $f\left(\cdot\right)$ to encode the re-occurrence features of historical neighbors by:
\begin{equation}
    \label{re-occur}
    \mathbf{X}_{i,\mathcal{R}}\left(t\right)=f\left(\mathcal{R}_i\left(t\right)\right) \in \mathbb{R}^{|\mathcal{N}_i\left(t\right)|\times d},
\end{equation}
where $f\left(\cdot\right)$ is a three-layer perceptron with ReLU activation, whose input and output dimensions are 1 and $d$, respectively.

For node $i$ at time $t$, we compute the embedding $\mathbf{z}_i(t)$ with its long-term memory $\mathcal{M}^L_i(t)$. We aggregate its historical neighbors’ long-term memory, $\mathcal{M}^L_j(t_j)$ where $j \in \mathcal{N}_i\left(t\right)$, using an attention mechanism as follows:
\begin{equation}
    \label{graph1}
    h_i^l(t) = \operatorname{MLP}^{(l)}\left(\mathbf{h}_i^{l-1}\left(t\right)\| \tilde{\mathbf{h}}_i^{l}\left(t\right)\right),
\end{equation}
\begin{equation}
    \label{graph2}
    \tilde{\mathbf{h}}_i^{l}\left(t\right) = \operatorname{Att}^{(l)}\left(\bigodot_{j \in \mathcal{N}_i\left(t\right)}\left( \mathbf{h}_j^{l-1}\left(t\right) \| \mathbf{e}_{ij}\left(t_j\right) \| \Phi\left(t - t_j\right) \| \mathbf{X}_{j,\mathcal{R}}\left(t_j\right)\right)\right)
\end{equation}
where $\bigodot$ denotes the stacking operation and $\operatorname{Att}(\cdot)$ is the graph attention used in \cite{TGN2020}. Note that the input $\mathbf{h}_i^0(t) = \mathcal{M}^L_i(t)$ and the node representation $\mathbf{z}_{i}(t) = \mathbf{h}_i^{L}(t)$ where $L$ is the layer number.

\subsection{Training}
\subsubsection{Error Grandients} Our method is differential except for the Bernoulli process in Equation \ref{bernoulli}, which is a binary value as the output. We employ the widely-used straight-through estimator \cite{skiprnn2017}, which implements the identity to approximate the step function for gradients computation during the backward pass: $\frac{\partial \operatorname{Bernoulli}(x)}{\partial x}=1.$

\subsubsection{Loss Function} We take temporal link prediction as our self-supervised task. For the representation of nodes $i$ and $j$ at time $t$, $\mathbf{z}_i(t)$ and $\mathbf{z}_j(t)$, we compute the probability of having interaction between them by a two-layer MLP:
\begin{equation}
    \label{loss1}
    \hat{p}_{ij}(t) = \sigma \left(\operatorname{MLP}\left(\mathbf{z}_i\left(t\right) \| \mathbf{z}_j\left(t\right)\right)\right),
\end{equation}
where $\sigma(\cdot)$ is the sigmoid function. Then, we set the cross-entropy as the loss function:
\begin{equation}
    \label{loss2}
    \mathcal{L}=-\sum_{(i, j, t) \in \mathcal{E}}\left[\log \hat{p}_{i j}(t)+\log \left(1-\hat{p}_{i k}(t)\right)\right],
\end{equation}
where $k$ is the negative destination node by random sampling. The pseudo-code of the {\our} is provided in Algorithm~\ref{algorithm}.

\section{Experiments}
\subsection{Datasets and Baselines}
\label{sec:ExpSetting}
For better comparison, we conduct experiments with four widely-used public datasets \cite{JODIE2019} including Wikipedia, Reddit, MOOC, and LastFM. Notably, all datasets have no node feature, and MOOC and LastFM have no edge feature, where we assign zero vectors in each of these datasets. Except for LastFM, others share evolving node labels of source nodes, and we can conduct the node classification task on them. All datasets are split with 70\%-15\%-15\% for training, validation, and testing as \cite{TGN2020}.

For evaluation, we choose nine dynamic graph modeling methods to compare with ours, including CTDNE \cite{CTDNE2018}, DyRep \cite{DyRep2019}, JODIE \cite{JODIE2019}, TGAT \cite{TGAT2020}, TGN \cite{TGN2020}, CAW \cite{caw2021}, TIGER \cite{tiger2023}, GraMixer \cite{graphmixer2023}, PINT \cite{pint2022}. Note that CTDNE can not be applied in the inductive setting, and CAW can not be conducted in the evolving node classification task.

\subsection{Temporal Link Prediction}
\label{sec:linkprediction}
Firstly, we evaluate our model on the temporal link prediction task. Similar to previous dynamic graph modeling methods, we test our model under two settings: transductive and inductive. In the transductive setting, we test edges whose nodes have been seen in the training splits, while in the inductive setting, we examine the unseen nodes for temporal link prediction. We use average precision (AP) as our evaluation metric and select an equal number of negative edges as we did in Equation \ref{loss2}. 

The results are shown in Table \ref{linkprediction}. Our model outperforms all baselines on all datasets in both transductive and inductive settings. This observation proves the excellent effectiveness and expressiveness of our method. For all baselines, existing sequential models, \textit{i.e.}, \cite{CTDNE2018, JODIE2019, DyRep2019, graphmixer2023}, perform worse than graph models. This may be owing to the fact that graph models, whose nodes can attend to multi-hop neighbors, have preserved the longer neighbors' information during training. It also gives us the motivation that there is still considerable room for improvement in sequential models.

\begin{table*}[t]
  \small
  \caption{\centering P-value of the chi-square independence test on Wikipedia and MOOC.}
  \label{p-value}
  \begin{tabular}{cccccccccc}
    \toprule
    & \textbf{JODIE} & \textbf{TGAT} & \textbf{DyRep} & \textbf{TGN} & \textbf{CAW} & \textbf{TIGER} & \textbf{GraMixer} & \textbf{PINT} & \textbf{Ours} \\
    \midrule
    \textbf{Wikipedia} & 0.006 & 0.015 & 0.011 & 0.031 & 0.008 & 0.042 & 0.029 & 0.040 & \textbf{0.185}  \\
    \textbf{MOOC}      & 0.010 & 0.005 & 0.018 & 0.034 & 0.019 & 0.010 & 0.033 & 0.022 & \textbf{0.115} \\
    \bottomrule
  \end{tabular}
\end{table*}
\begin{figure*}
  \centering
  \begin{subfigure}{.46\linewidth}
    \centering
    \includegraphics[width=\linewidth]{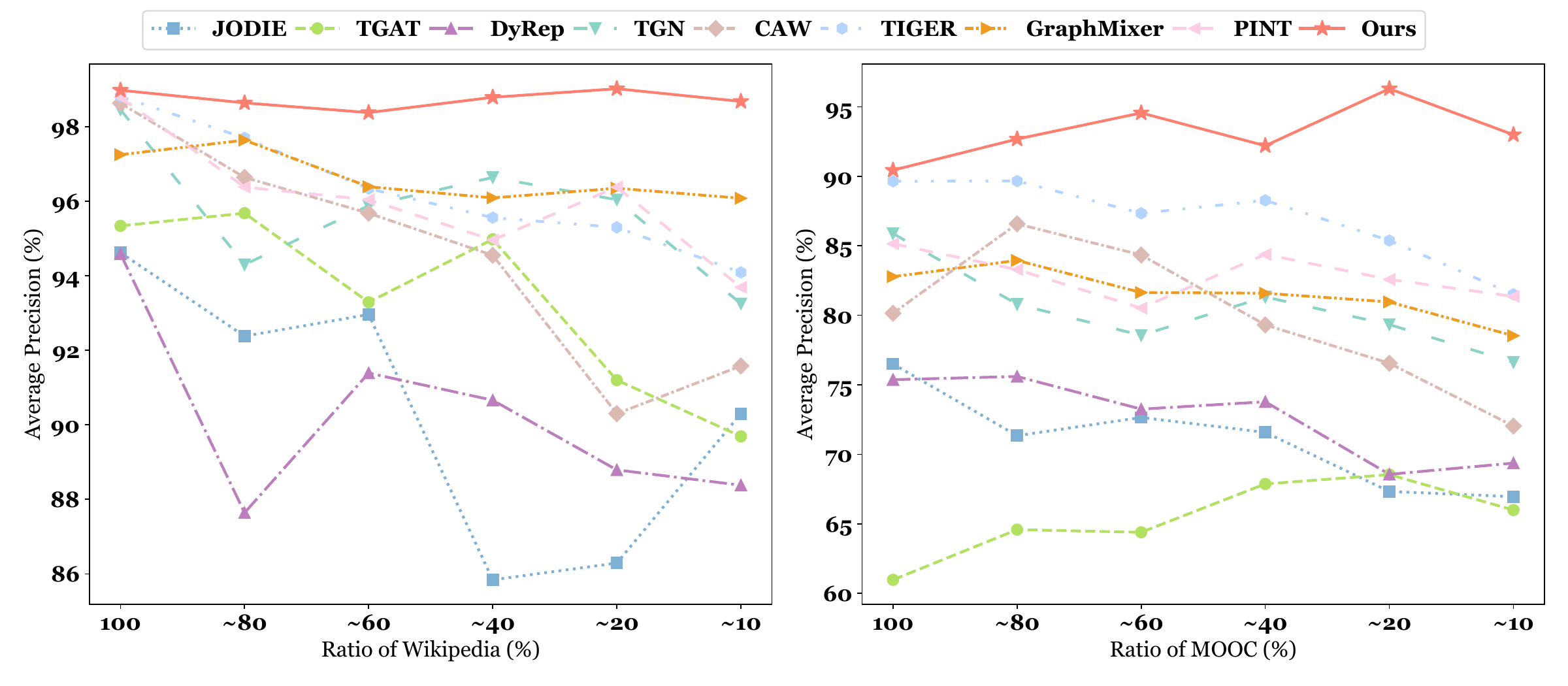}
    \caption{Transductive temporal link prediction.}
    \label{long-term-trans}
  \end{subfigure}
  \centering
  \begin{subfigure}{.46\linewidth}
    \centering
    \includegraphics[width=\linewidth]{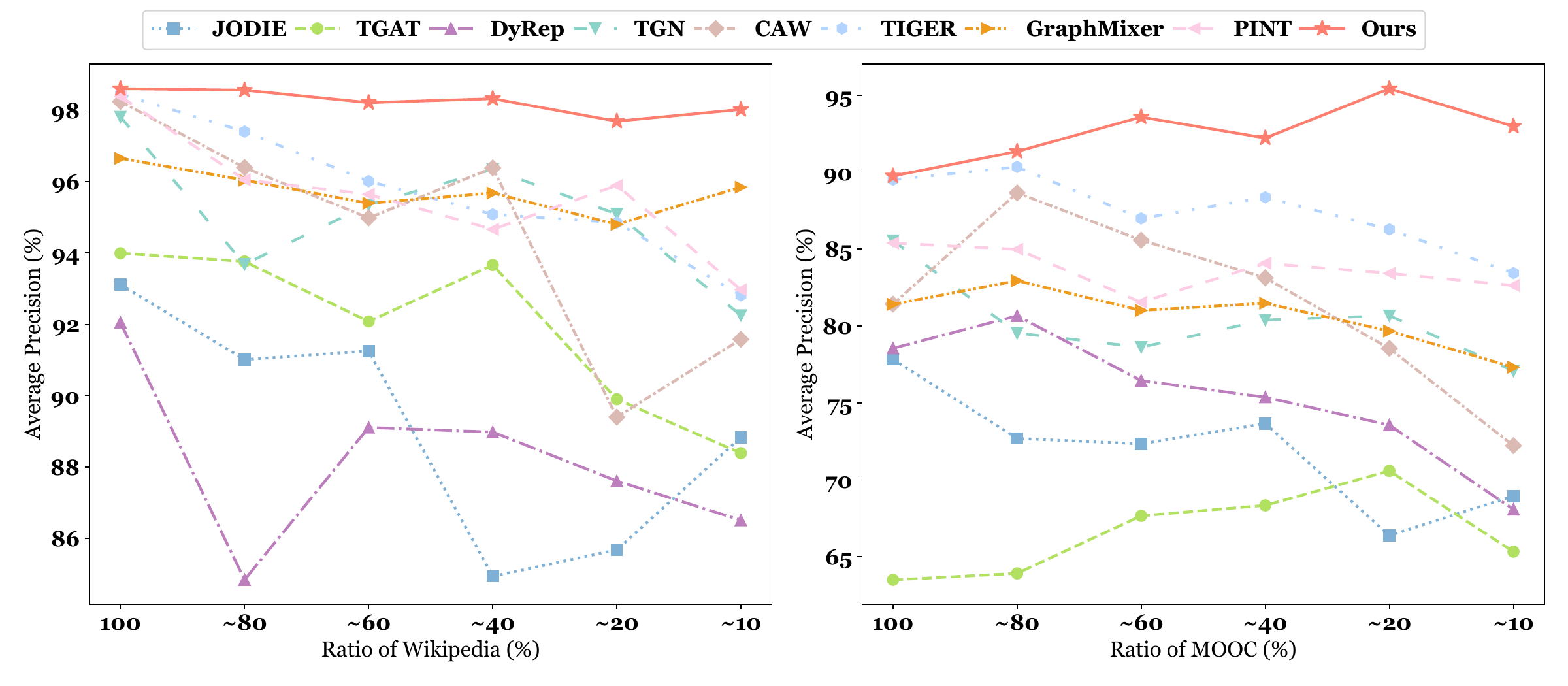}
    \caption{Inductive temporal link prediction.}
    \label{long-term-in}
  \end{subfigure}
  \caption{The ability to node-wise long-term modeling for temporal link prediction task on Wikipedia and MOOC.}
  \label{long-term}
\end{figure*}

~\subsection{Evolving Node Classification}
\label{sec:nodeclassification}
To further evaluate the effectiveness of our model, we use the learned temporal representation for the evolving node classification task. In practice, we utilize temporal link prediction as a pre-training task for the models. We use Wikipedia, Reddit, and MOOC for testing as only these datasets have evolving node labels. Following \cite{TGN2020}, we input the temporal representation of node $i$, $\mathbf{z}_i(t)$, into a two-layer MLP to obtain the class probability of the temporal nodes and then design a training signal in Equation \ref{loss2}. 

The results are presented in Table \ref{evolvingClassification}. Our method achieves the best performance on all datasets, further confirming the powerful dynamic graph modeling capabilities of our method. The satisfactory outcomes demonstrate that the learned representations of our method are effective for downstream tasks. 
\begin{figure*}
  \centering
  \begin{subfigure}{.3\linewidth}
    \centering
    \includegraphics[width=\linewidth]{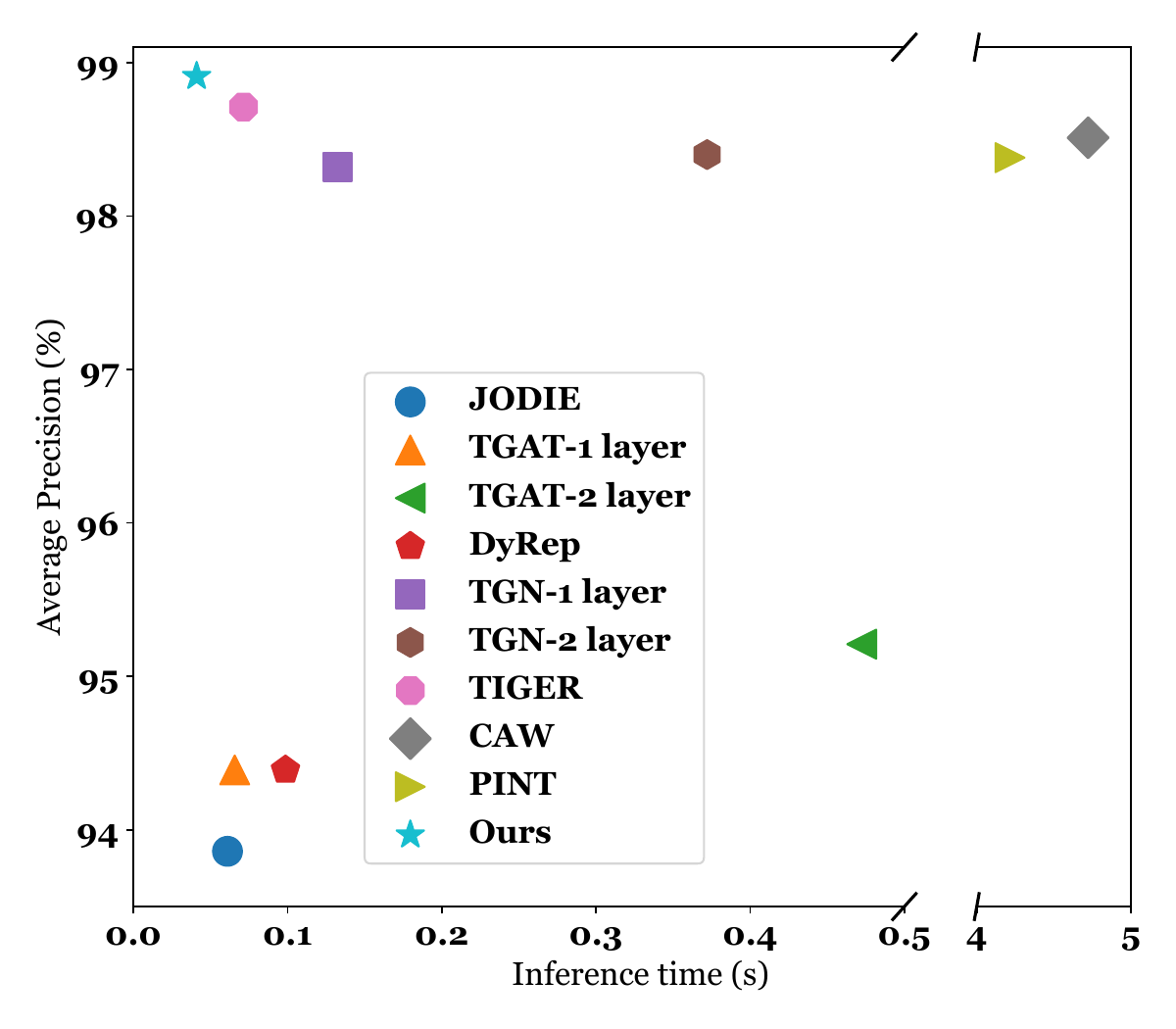}
    \caption{Wikipedia.}
    \label{infer_wiki}
  \end{subfigure}
  \centering
  \begin{subfigure}{.3\linewidth}
    \centering
    \includegraphics[width=\linewidth]{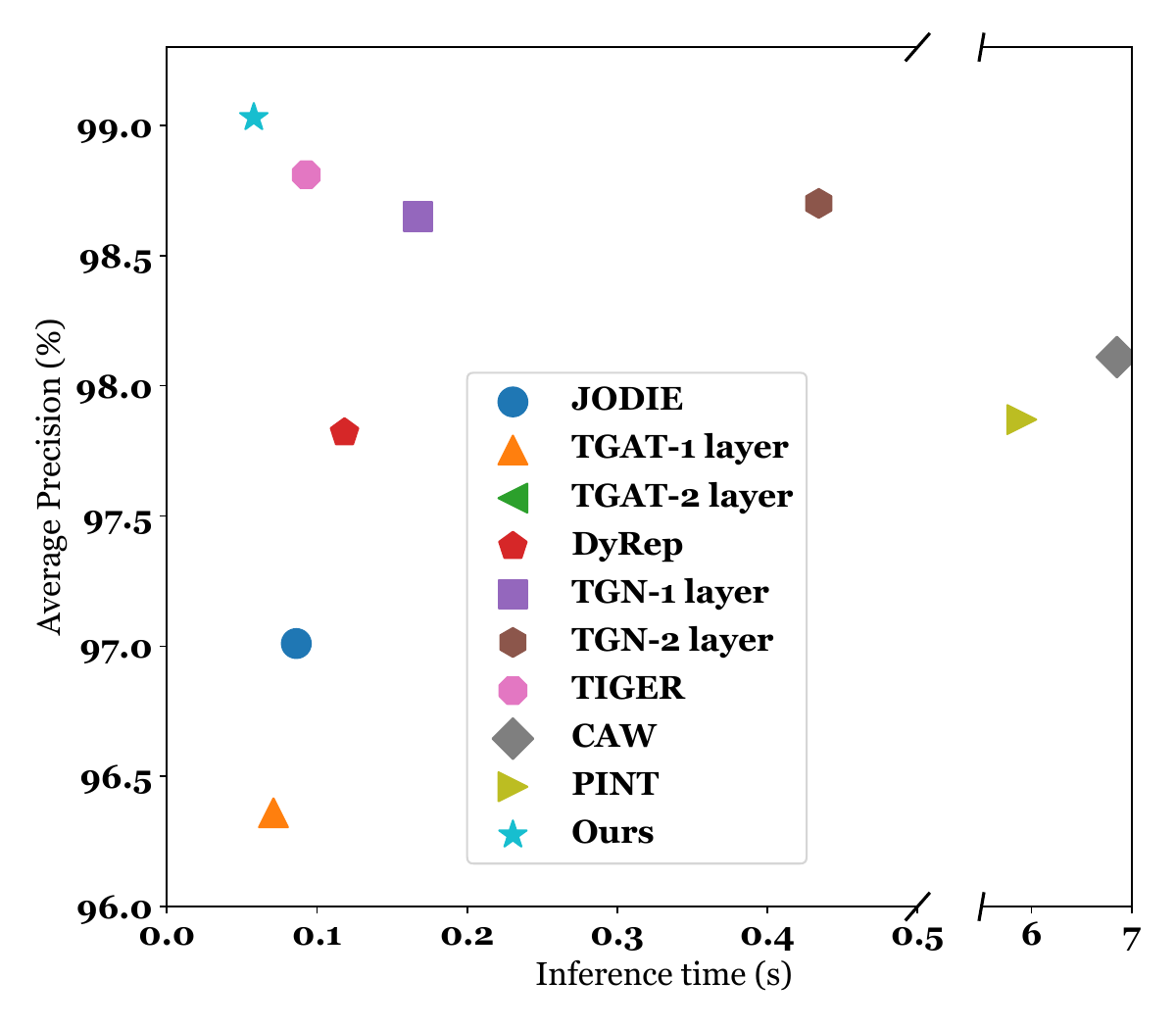}
    \caption{Reddit.}
    \label{infer_reddit}
  \end{subfigure}
  \centering
  \begin{subfigure}{.3\linewidth}
    \centering
    \includegraphics[width=\linewidth]{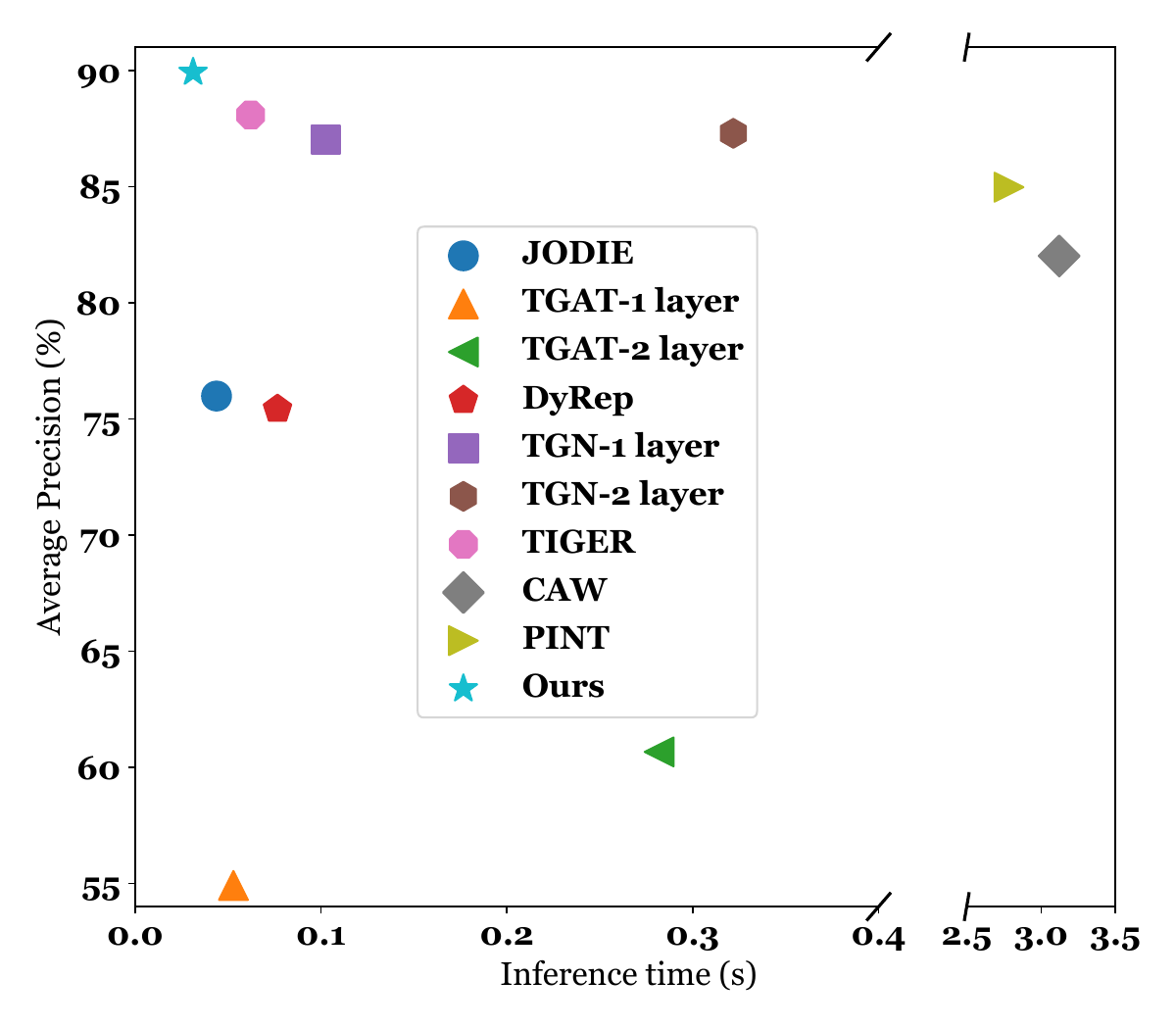}
    \caption{MOOC.}
    \label{infer_mooc}
  \end{subfigure}
  \caption{Analysis between the inference time of a batch and the performance in transductive temporal link prediction task.}
  \label{infer}
\end{figure*}

\subsection{Ability to Node-wise Long-term Modeling}
To validate the node-wise long-term modeling ability of models, we design experiments focusing on big nodes in dynamic graphs. Specifically, we sort all the nodes in dynamic graphs by the number of their edges, \textit{i.e.}, node frequency, and select nodes whose node frequency is in the top $k \in \{100\%, 80\%, 60\%, 40\%, 20\%, 10\%\}$ to generate some subgraphs separately. It is worth noting that the smaller $k$ of the subgraph, the higher proportion of big nodes, the more challenging node-wise long-term modeling. Moreover, considering that the unequal number of samples in these subgraphs may affect the credibility of the conclusion, we employ the chi-square independence test. We first conduct a contingency table with the number of successful and failed predictions that are generated from the model in each subgraph, then we calculate the P-value of the chi-square independence test $p_v$. Our null hypothesis is ``the subgraphs and the success or failure of the predictions are independent''. If $p_v>0.05$, we can accept the null hypothesis. It indicates that different subgraphs, which contain different proportions of big nodes, have little impact on the performance of the model, confirming the model's ability to node-wise long-term modeling. 

As shown in Figure \ref{long-term} and Table \ref{p-value}, Our model outperforms all other baselines in all subgraphs. As the proportion of big nodes increases, the performance of our model remains stable, while other baselines decline, demonstrating the strong node-wise long-term modeling capability of our model in dynamic graphs. Only our model has a P-value greater than 0.05. The chi-squared test rules out concerns that may have arisen from differences in sample number among subgraphs, enhancing the credibility of the conclusion.

\subsection{Analysis of Inference Time}
To verify the effectiveness of discarding edges and the model's efficiency, we conduct comparative experiments on the inference time and the performance of models. Our experiments are performed on a Linux PC with an Intel i7 CPU (6 cores, 2.6 GHz), using the original public implementations of baselines. In industry, the inference time of a model is much more important than its training time. In the online payment platform, for example, there are billions of transaction data that are generated daily. Industry research institutes do not necessarily train models frequently, but they need to process these large amounts of daily data frequently for downstream tasks such as financial risk management \cite{apan2021}, leading to redundant time consumption. Consequently, a model with a lower inference time has more commercial value. Thus, we compare the inference time of a batch (batch size is 100) and the performance of models. 

The results are shown in Figure \ref{infer}, where the closer to the upper left corner, the shorter the inference time and the better performance of the model. Our model outperforms other baselines in both inference time and performance, mainly due to the successful removal of some useless or noisy edges. We also find that \cite{caw2021, pint2022} have significantly longer inference time compared to other methods. This may be because they search for neighbors through temporal walks, which is extremely time-consuming during inference.

\subsection{Ablation Study}
\label{sec:ablationstudy}
We conduct an ablation study to further investigate the impact of the main innovative components in our model, including the state module (SM) in Section \ref{sec:short-term}, Gaussian Range Encoding (GRE) in Section \ref{sec:Gaussian}, Identity Attention (IA) in Section \ref{sec:identity-attention}, and the Re-occurrence features (ReO) in Section \ref{sec:re-occurrence}. We propose four variants: w/o SM, w/o GRE, w/o IA, and w/o ReO, respectively. The w/o SM variant does not discard temporal edges and performs indiscriminate updating; The w/o GRE variant replaces the Gaussian Range Encoding with the default positional encoding used in Transformer \cite{transformer2017}; The w/o IA variant replaces Identity Attention with full attention \cite{transformer2017}; The w/o ReO variant removes the re-occurrence features from the graph module. 

\begin{figure}
  \centering
  \begin{subfigure}{.49\linewidth}
    \centering
    \includegraphics[width=\linewidth]{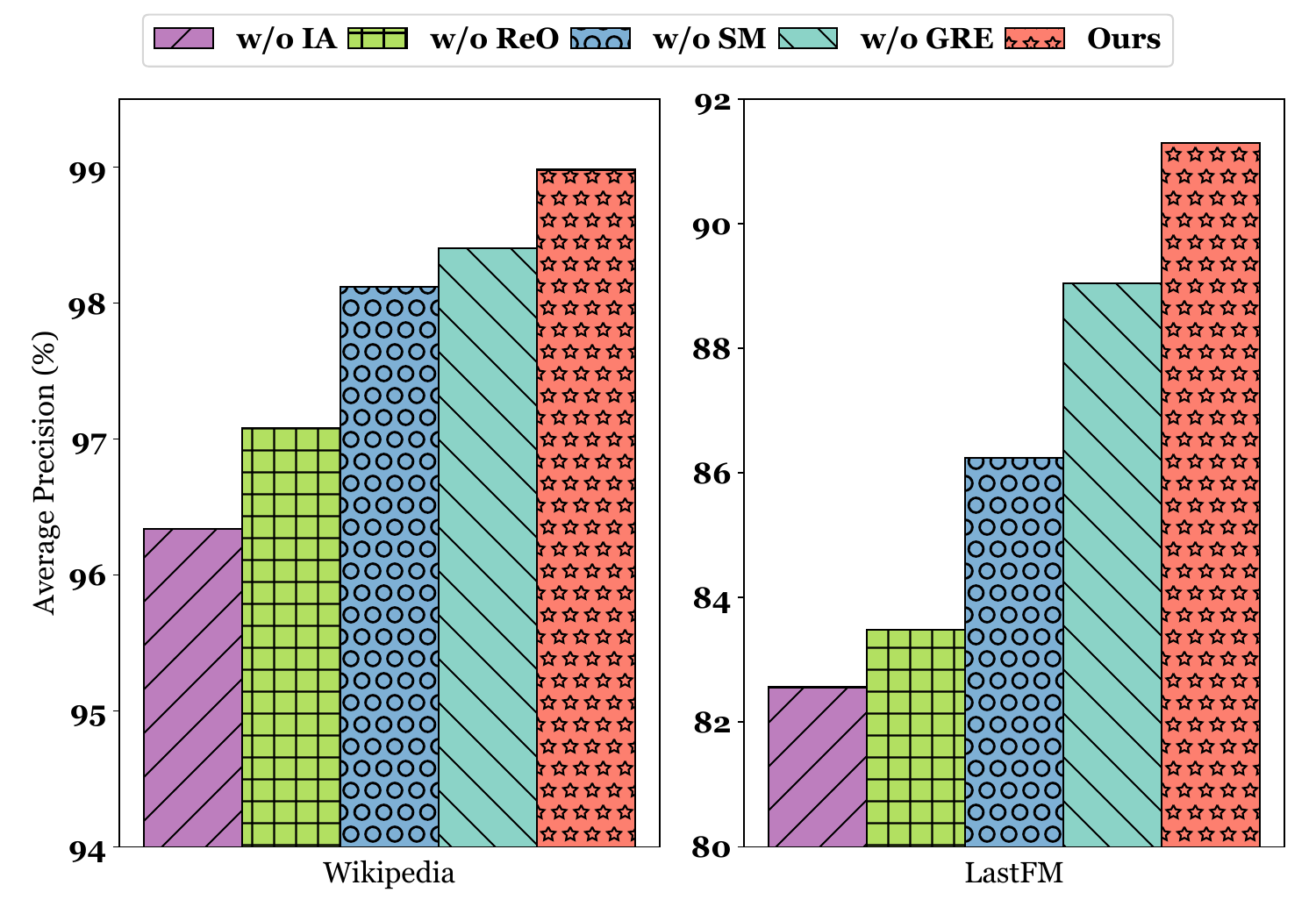}
    \caption{Transductive setting.}
    \label{ablation-trans}
  \end{subfigure}
  \centering
  \begin{subfigure}{.49\linewidth}
    \centering
    \includegraphics[width=\linewidth]{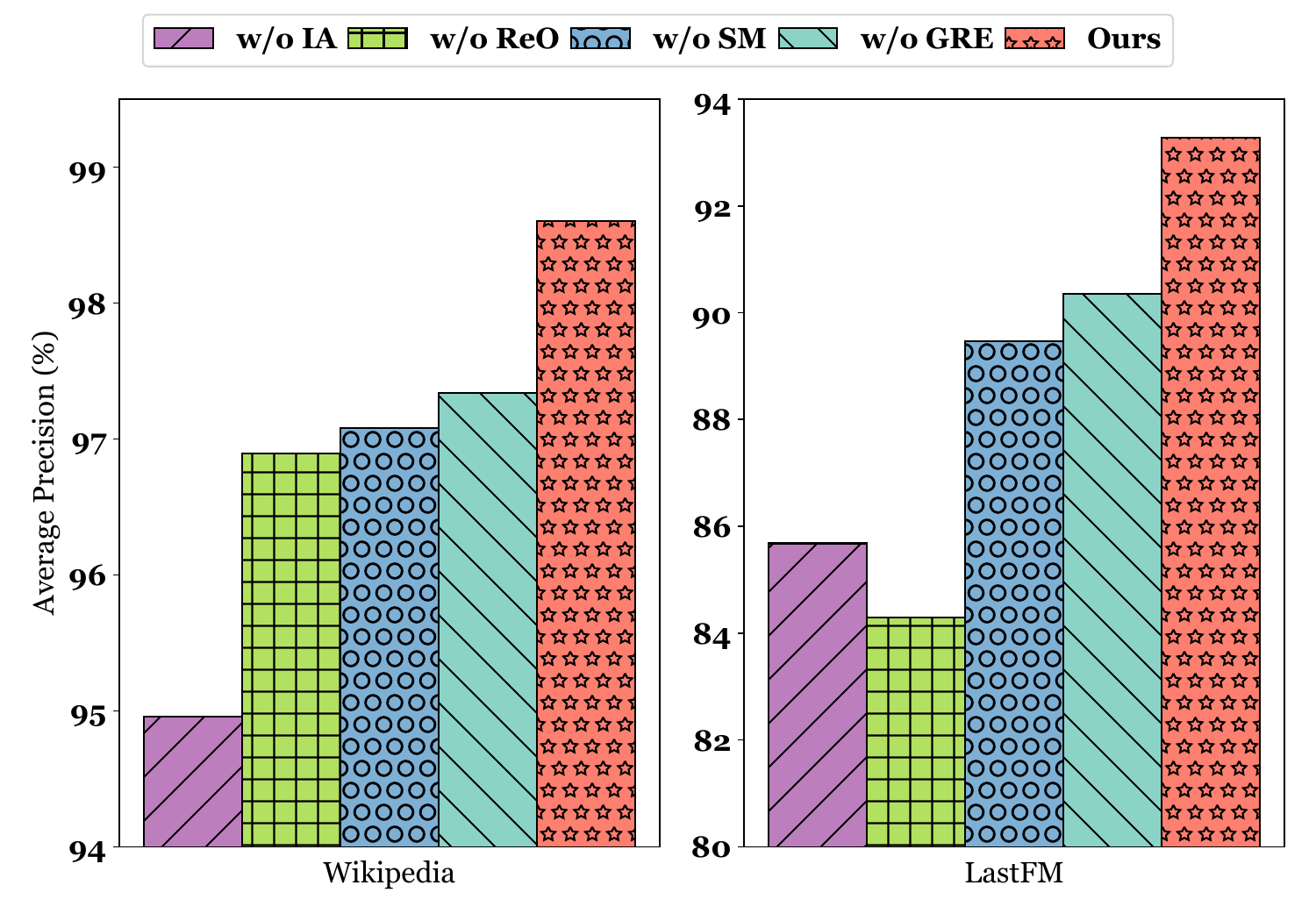}
    \caption{Inductive setting.}
    \label{ablation-in}
  \end{subfigure}
  \caption{Ablation study for transductive and inductive temporal link prediction task on Wikipedia and LastFM.}
  \label{ablation}
\end{figure}

We report the results for link prediction on Wikipedia and LastFM, as shown in figure \ref{ablation}. Our model achieves the best performance when using all components, and the performance decreases when each component is removed or replaced with the default one. The Identity Attention and re-occurrence features have the most significant impact on the model, indicating that they may be able to better extract valuable information in dynamic graphs.

\subsection{Parameter Study}
We conduct a parameter study to better investigate the impact of main hyper-parameters in our model, including the block number of Transformer $b$ in Section \ref{sec:transformer}, the memory dimension $d$ in Equation \ref{long-term-memory}, the chunk size or window length for long-term modeling $n$ in Equation \ref{attention4}, and the window size for short-term modeling $s$ in Section \ref{sec:short-long}. We conduct experiments on lastFM, and we find that increasing the value of $b$ does not lead to better performance and the best cost-effectiveness is achieved when $b=$ 2 or 3. Through our analysis of $n$, our model shows improved performance in modeling longer sequences, indicating that our model is able to effectively capture longer dependencies. Other hyper-parameters have varying degrees of impact on our results.
\begin{figure}
  \centering
  \begin{subfigure}{0.4\linewidth}
    \centering
    \includegraphics[width=\linewidth]{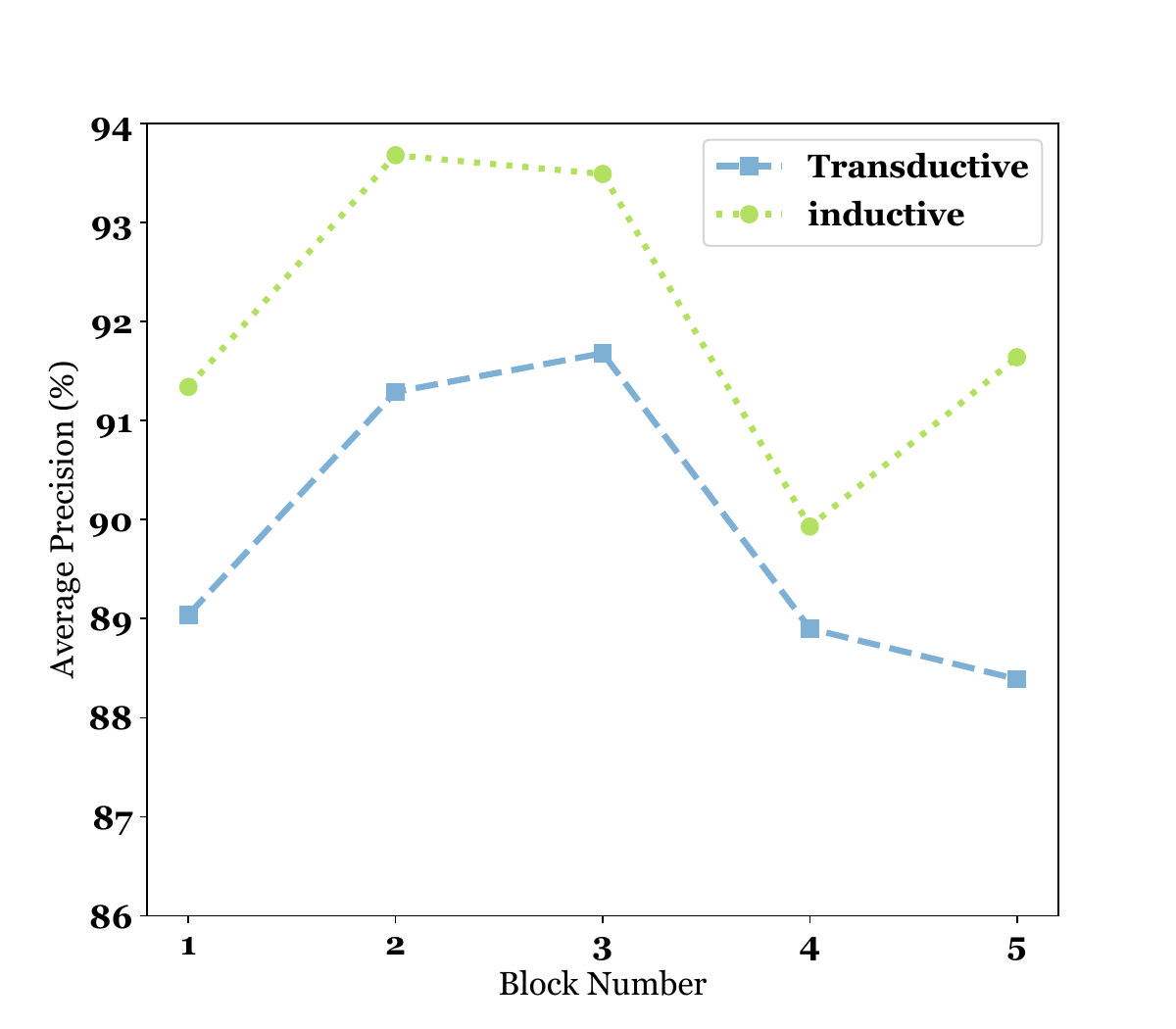}
    \caption{Block number $b$.}
    \label{block}
  \end{subfigure}
  \centering
  \begin{subfigure}{0.4\linewidth}
    \centering
    \includegraphics[width=\linewidth]{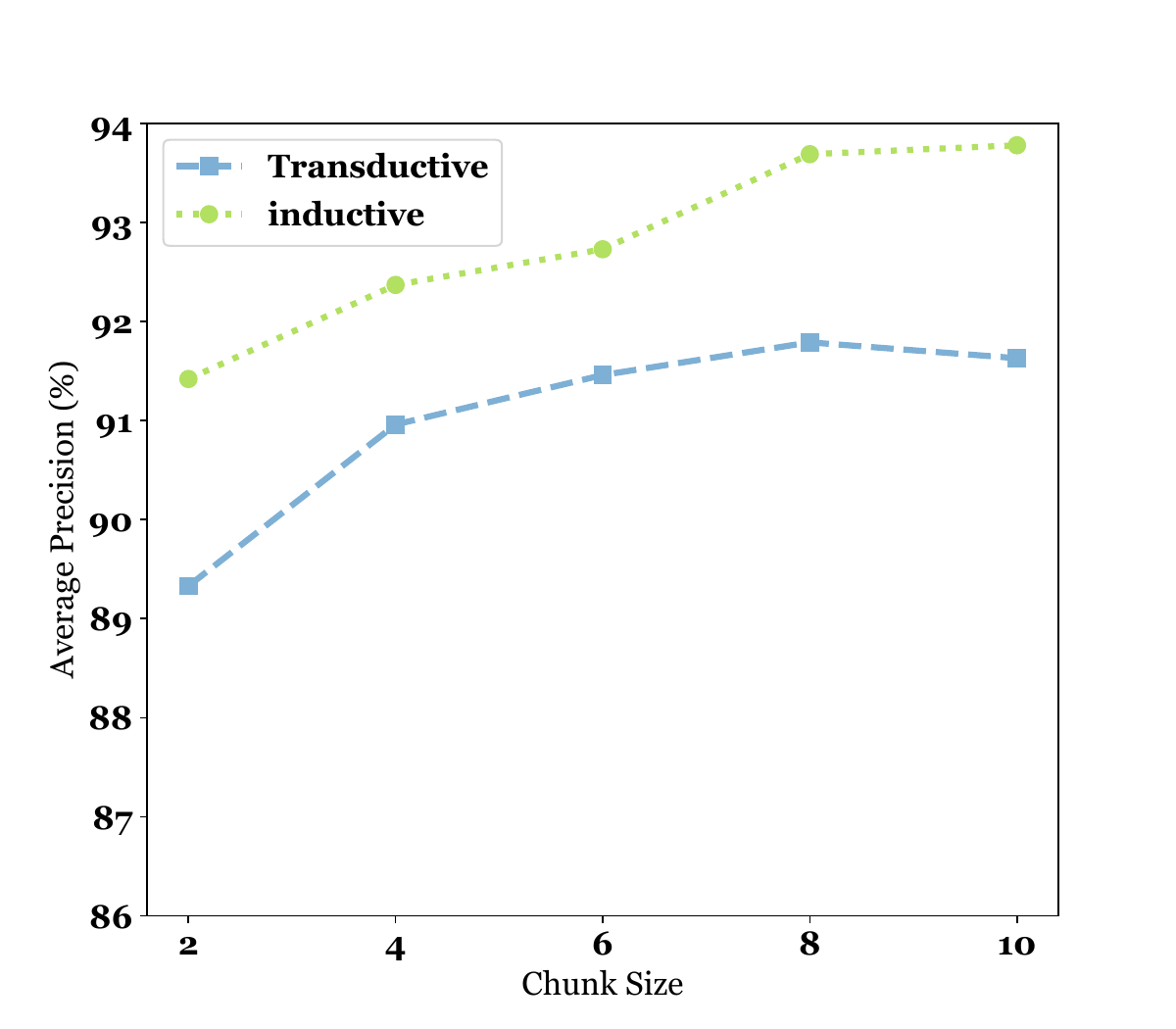}
    \caption{Chunk size $n$.}
    \label{h}
  \end{subfigure}
  \centering
  \begin{subfigure}{0.4\linewidth}
    \centering
    \includegraphics[width=\linewidth]{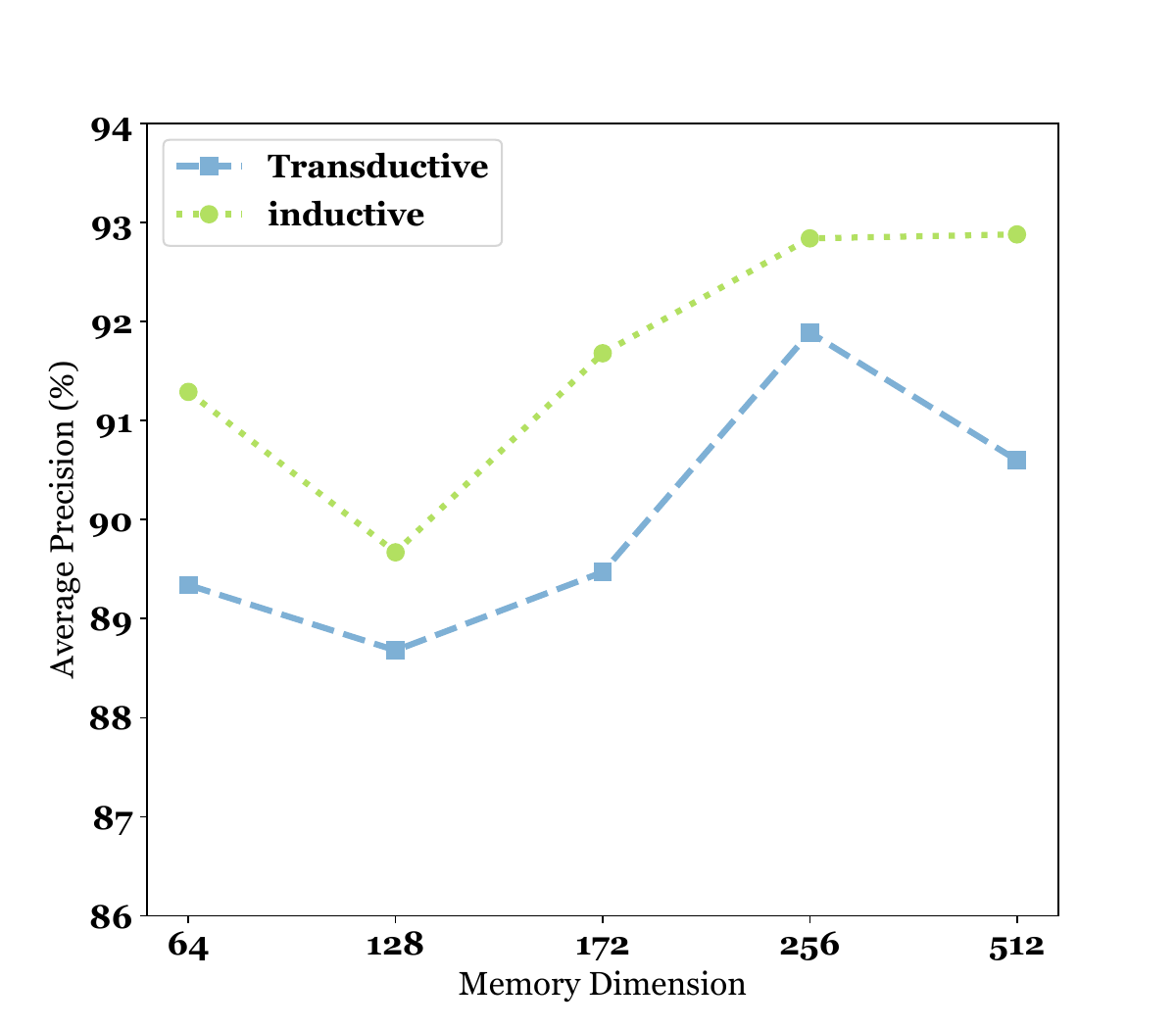}
    \caption{Memory dimension $d$.}
    \label{block}
  \end{subfigure}
  \centering
  \begin{subfigure}{0.4\linewidth}
    \centering
    \includegraphics[width=\linewidth]{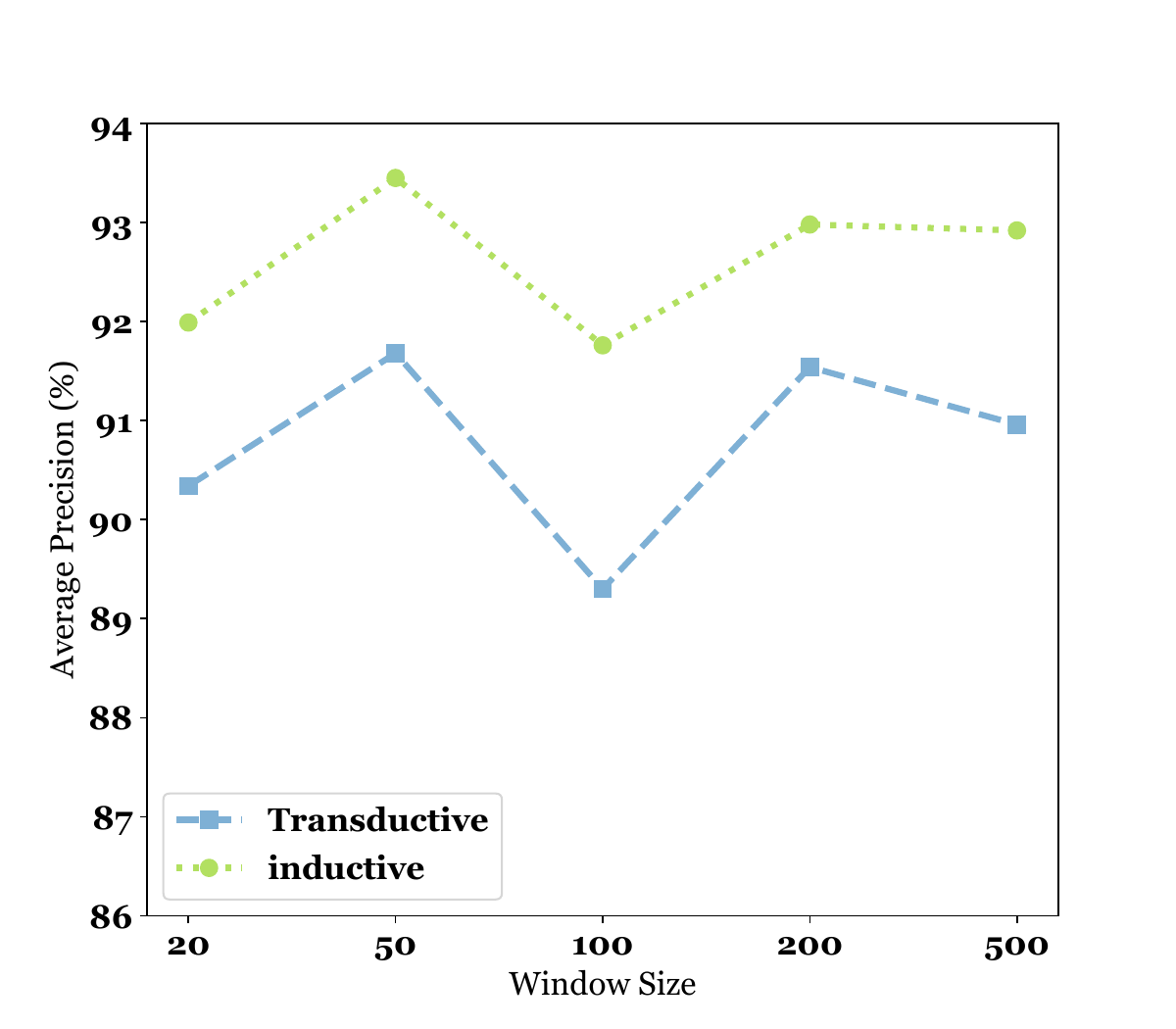}
    \caption{Window size $s$.}
    \label{h}
  \end{subfigure}
  \caption{Parameter study in transductive and inductive temporal link prediction task on LastFM.}
  \label{sensitivity}
\end{figure}

\section{Conclusion and Future Work}
In this paper, we propose {\our}, a dynamic graph modeling method with instant long-term modeling and re-occurrence preservation. We introduce a state module to enhance inference efficiency and prevent noisy information. We further propose Identity Attention to empower a Transformer-based updater for long-term modeling and successfully encode re-occurrence features into the graph module. For future work, we hope to design a dynamic graph modeling method based entirely on Transformer architecture in the future.

\begin{acks}
This work is funded in part by the National Natural Science Foundation of China Projects No. U1936213, No. 62206059, the Shanghai Science and Technology Development Fund No.22dz1200704, NSF through grants IIS-1763365 and IIS-2106972, and also supported by CNKLSTISS.
\end{acks}


\bibliographystyle{ACM-Reference-Format}
\balance
\bibliography{sample-base}


\begin{thebibliography}{38}


\ifx \showCODEN    \undefined \def \showCODEN     #1{\unskip}     \fi
\ifx \showDOI      \undefined \def \showDOI       #1{#1}\fi
\ifx \showISBNx    \undefined \def \showISBNx     #1{\unskip}     \fi
\ifx \showISBNxiii \undefined \def \showISBNxiii  #1{\unskip}     \fi
\ifx \showISSN     \undefined \def \showISSN      #1{\unskip}     \fi
\ifx \showLCCN     \undefined \def \showLCCN      #1{\unskip}     \fi
\ifx \shownote     \undefined \def \shownote      #1{#1}          \fi
\ifx \showarticletitle \undefined \def \showarticletitle #1{#1}   \fi
\ifx \showURL      \undefined \def \showURL       {\relax}        \fi
\providecommand\bibfield[2]{#2}
\providecommand\bibinfo[2]{#2}
\providecommand\natexlab[1]{#1}
\providecommand\showeprint[2][]{arXiv:#2}

\bibitem[\protect\citeauthoryear{Battaglia, Hamrick, Bapst, Sanchez-Gonzalez,
  Zambaldi, Malinowski, Tacchetti, Raposo, Santoro, Faulkner,
  et~al\mbox{.}}{Battaglia et~al\mbox{.}}{2018}]%
        {nlp3_2018}
\bibfield{author}{\bibinfo{person}{Peter~W Battaglia},
  \bibinfo{person}{Jessica~B Hamrick}, \bibinfo{person}{Victor Bapst},
  \bibinfo{person}{Alvaro Sanchez-Gonzalez}, \bibinfo{person}{Vinicius
  Zambaldi}, \bibinfo{person}{Mateusz Malinowski}, \bibinfo{person}{Andrea
  Tacchetti}, \bibinfo{person}{David Raposo}, \bibinfo{person}{Adam Santoro},
  \bibinfo{person}{Ryan Faulkner}, {et~al\mbox{.}}}
  \bibinfo{year}{2018}\natexlab{}.
\newblock \showarticletitle{Relational inductive biases, deep learning, and
  graph networks}.
\newblock \bibinfo{journal}{\emph{arXiv preprint arXiv:1806.01261}}
  (\bibinfo{year}{2018}).
\newblock


\bibitem[\protect\citeauthoryear{Campos, Jou, Gir{\'o}-i Nieto, Torres, and
  Chang}{Campos et~al\mbox{.}}{2017}]%
        {skiprnn2017}
\bibfield{author}{\bibinfo{person}{V{\'\i}ctor Campos},
  \bibinfo{person}{Brendan Jou}, \bibinfo{person}{Xavier Gir{\'o}-i Nieto},
  \bibinfo{person}{Jordi Torres}, {and} \bibinfo{person}{Shih-Fu Chang}.}
  \bibinfo{year}{2017}\natexlab{}.
\newblock \showarticletitle{Skip rnn: Learning to skip state updates in
  recurrent neural networks}.
\newblock \bibinfo{journal}{\emph{arXiv preprint arXiv:1708.06834}}
  (\bibinfo{year}{2017}).
\newblock


\bibitem[\protect\citeauthoryear{Carion, Massa, Synnaeve, Usunier, Kirillov,
  and Zagoruyko}{Carion et~al\mbox{.}}{2020}]%
        {cv1_2020}
\bibfield{author}{\bibinfo{person}{Nicolas Carion}, \bibinfo{person}{Francisco
  Massa}, \bibinfo{person}{Gabriel Synnaeve}, \bibinfo{person}{Nicolas
  Usunier}, \bibinfo{person}{Alexander Kirillov}, {and} \bibinfo{person}{Sergey
  Zagoruyko}.} \bibinfo{year}{2020}\natexlab{}.
\newblock \showarticletitle{End-to-end object detection with transformers}. In
  \bibinfo{booktitle}{\emph{Computer Vision--ECCV 2020: 16th European
  Conference, Glasgow, UK, August 23--28, 2020, Proceedings, Part I 16}}.
  Springer, \bibinfo{pages}{213--229}.
\newblock


\bibitem[\protect\citeauthoryear{Chen, Zhu, Xu, Liu, Xiong, Zhang, and
  Song}{Chen et~al\mbox{.}}{2021}]%
        {edge2021}
\bibfield{author}{\bibinfo{person}{Xinshi Chen}, \bibinfo{person}{Yan Zhu},
  \bibinfo{person}{Haowen Xu}, \bibinfo{person}{Mengyang Liu},
  \bibinfo{person}{Liang Xiong}, \bibinfo{person}{Muhan Zhang}, {and}
  \bibinfo{person}{Le Song}.} \bibinfo{year}{2021}\natexlab{}.
\newblock \showarticletitle{Efficient Dynamic Graph Representation Learning at
  Scale}.
\newblock \bibinfo{journal}{\emph{arXiv preprint arXiv:2112.07768}}
  (\bibinfo{year}{2021}).
\newblock


\bibitem[\protect\citeauthoryear{Cho, Van~Merri{\"e}nboer, Gulcehre, Bahdanau,
  Bougares, Schwenk, and Bengio}{Cho et~al\mbox{.}}{2014}]%
        {GRU2014}
\bibfield{author}{\bibinfo{person}{Kyunghyun Cho}, \bibinfo{person}{Bart
  Van~Merri{\"e}nboer}, \bibinfo{person}{Caglar Gulcehre},
  \bibinfo{person}{Dzmitry Bahdanau}, \bibinfo{person}{Fethi Bougares},
  \bibinfo{person}{Holger Schwenk}, {and} \bibinfo{person}{Yoshua Bengio}.}
  \bibinfo{year}{2014}\natexlab{}.
\newblock \showarticletitle{Learning phrase representations using RNN
  encoder-decoder for statistical machine translation}.
\newblock \bibinfo{journal}{\emph{arXiv preprint arXiv:1406.1078}}
  (\bibinfo{year}{2014}).
\newblock


\bibitem[\protect\citeauthoryear{Cong, Zhang, Kang, Yuan, Wu, Zhou, Tong, and
  Mahdavi}{Cong et~al\mbox{.}}{2023}]%
        {graphmixer2023}
\bibfield{author}{\bibinfo{person}{Weilin Cong}, \bibinfo{person}{Si Zhang},
  \bibinfo{person}{Jian Kang}, \bibinfo{person}{Baichuan Yuan},
  \bibinfo{person}{Hao Wu}, \bibinfo{person}{Xin Zhou},
  \bibinfo{person}{Hanghang Tong}, {and} \bibinfo{person}{Mehrdad Mahdavi}.}
  \bibinfo{year}{2023}\natexlab{}.
\newblock \showarticletitle{Do We Really Need Complicated Model Architectures
  For Temporal Networks?}
\newblock \bibinfo{journal}{\emph{arXiv preprint arXiv:2302.11636}}
  (\bibinfo{year}{2023}).
\newblock


\bibitem[\protect\citeauthoryear{Dai, Wang, Trivedi, and Song}{Dai
  et~al\mbox{.}}{2016}]%
        {deepco2016}
\bibfield{author}{\bibinfo{person}{Hanjun Dai}, \bibinfo{person}{Yichen Wang},
  \bibinfo{person}{Rakshit Trivedi}, {and} \bibinfo{person}{Le Song}.}
  \bibinfo{year}{2016}\natexlab{}.
\newblock \showarticletitle{Deep coevolutionary network: Embedding user and
  item features for recommendation}.
\newblock \bibinfo{journal}{\emph{arXiv preprint arXiv:1609.03675}}
  (\bibinfo{year}{2016}).
\newblock


\bibitem[\protect\citeauthoryear{Devlin, Chang, Lee, and Toutanova}{Devlin
  et~al\mbox{.}}{2018}]%
        {nlp1_2018}
\bibfield{author}{\bibinfo{person}{Jacob Devlin}, \bibinfo{person}{Ming-Wei
  Chang}, \bibinfo{person}{Kenton Lee}, {and} \bibinfo{person}{Kristina
  Toutanova}.} \bibinfo{year}{2018}\natexlab{}.
\newblock \showarticletitle{Bert: Pre-training of deep bidirectional
  transformers for language understanding}.
\newblock \bibinfo{journal}{\emph{arXiv preprint arXiv:1810.04805}}
  (\bibinfo{year}{2018}).
\newblock


\bibitem[\protect\citeauthoryear{Dosovitskiy, Beyer, Kolesnikov, Weissenborn,
  Zhai, Unterthiner, Dehghani, Minderer, Heigold, Gelly,
  et~al\mbox{.}}{Dosovitskiy et~al\mbox{.}}{2020}]%
        {cv2_2020}
\bibfield{author}{\bibinfo{person}{Alexey Dosovitskiy}, \bibinfo{person}{Lucas
  Beyer}, \bibinfo{person}{Alexander Kolesnikov}, \bibinfo{person}{Dirk
  Weissenborn}, \bibinfo{person}{Xiaohua Zhai}, \bibinfo{person}{Thomas
  Unterthiner}, \bibinfo{person}{Mostafa Dehghani}, \bibinfo{person}{Matthias
  Minderer}, \bibinfo{person}{Georg Heigold}, \bibinfo{person}{Sylvain Gelly},
  {et~al\mbox{.}}} \bibinfo{year}{2020}\natexlab{}.
\newblock \showarticletitle{An image is worth 16x16 words: Transformers for
  image recognition at scale}.
\newblock \bibinfo{journal}{\emph{arXiv preprint arXiv:2010.11929}}
  (\bibinfo{year}{2020}).
\newblock


\bibitem[\protect\citeauthoryear{Dwivedi and Bresson}{Dwivedi and
  Bresson}{2020}]%
        {graph_transformer_layer2020}
\bibfield{author}{\bibinfo{person}{Vijay~Prakash Dwivedi} {and}
  \bibinfo{person}{Xavier Bresson}.} \bibinfo{year}{2020}\natexlab{}.
\newblock \showarticletitle{A generalization of transformer networks to
  graphs}.
\newblock \bibinfo{journal}{\emph{arXiv preprint arXiv:2012.09699}}
  (\bibinfo{year}{2020}).
\newblock


\bibitem[\protect\citeauthoryear{Kumar, Zhang, and Leskovec}{Kumar
  et~al\mbox{.}}{2019}]%
        {JODIE2019}
\bibfield{author}{\bibinfo{person}{Srijan Kumar}, \bibinfo{person}{Xikun
  Zhang}, {and} \bibinfo{person}{Jure Leskovec}.}
  \bibinfo{year}{2019}\natexlab{}.
\newblock \showarticletitle{Predicting dynamic embedding trajectory in temporal
  interaction networks}. In \bibinfo{booktitle}{\emph{Proceedings of the 25th
  ACM SIGKDD international conference on knowledge discovery \& data mining}}.
  \bibinfo{pages}{1269--1278}.
\newblock


\bibitem[\protect\citeauthoryear{Li, Cui, Wang, Zhang, Chen, and Wu}{Li
  et~al\mbox{.}}{2021}]%
        {that2021}
\bibfield{author}{\bibinfo{person}{Bing Li}, \bibinfo{person}{Wei Cui},
  \bibinfo{person}{Wei Wang}, \bibinfo{person}{Le Zhang},
  \bibinfo{person}{Zhenghua Chen}, {and} \bibinfo{person}{Min Wu}.}
  \bibinfo{year}{2021}\natexlab{}.
\newblock \showarticletitle{Two-stream convolution augmented transformer for
  human activity recognition}. In \bibinfo{booktitle}{\emph{Proceedings of the
  AAAI Conference on Artificial Intelligence}}, Vol.~\bibinfo{volume}{35}.
  \bibinfo{pages}{286--293}.
\newblock


\bibitem[\protect\citeauthoryear{Li, Jin, Xuan, Zhou, Chen, Wang, and Yan}{Li
  et~al\mbox{.}}{2019}]%
        {ts1_2019}
\bibfield{author}{\bibinfo{person}{Shiyang Li}, \bibinfo{person}{Xiaoyong Jin},
  \bibinfo{person}{Yao Xuan}, \bibinfo{person}{Xiyou Zhou},
  \bibinfo{person}{Wenhu Chen}, \bibinfo{person}{Yu-Xiang Wang}, {and}
  \bibinfo{person}{Xifeng Yan}.} \bibinfo{year}{2019}\natexlab{}.
\newblock \showarticletitle{Enhancing the locality and breaking the memory
  bottleneck of transformer on time series forecasting}.
\newblock \bibinfo{journal}{\emph{Advances in neural information processing
  systems}}  \bibinfo{volume}{32} (\bibinfo{year}{2019}).
\newblock


\bibitem[\protect\citeauthoryear{Liu, Zhao, Su, Cen, Qiu, Zhang, Wu, Dong, and
  Tang}{Liu et~al\mbox{.}}{2022}]%
        {gtransformer2_2022}
\bibfield{author}{\bibinfo{person}{Xiao Liu}, \bibinfo{person}{Shiyu Zhao},
  \bibinfo{person}{Kai Su}, \bibinfo{person}{Yukuo Cen},
  \bibinfo{person}{Jiezhong Qiu}, \bibinfo{person}{Mengdi Zhang},
  \bibinfo{person}{Wei Wu}, \bibinfo{person}{Yuxiao Dong}, {and}
  \bibinfo{person}{Jie Tang}.} \bibinfo{year}{2022}\natexlab{}.
\newblock \showarticletitle{Mask and Reason: Pre-Training Knowledge Graph
  Transformers for Complex Logical Queries}. In
  \bibinfo{booktitle}{\emph{Proceedings of the 28th ACM SIGKDD Conference on
  Knowledge Discovery and Data Mining}}. \bibinfo{pages}{1120--1130}.
\newblock


\bibitem[\protect\citeauthoryear{Liu, Ott, Goyal, Du, Joshi, Chen, Levy, Lewis,
  Zettlemoyer, and Stoyanov}{Liu et~al\mbox{.}}{2019}]%
        {nlp2_2019}
\bibfield{author}{\bibinfo{person}{Yinhan Liu}, \bibinfo{person}{Myle Ott},
  \bibinfo{person}{Naman Goyal}, \bibinfo{person}{Jingfei Du},
  \bibinfo{person}{Mandar Joshi}, \bibinfo{person}{Danqi Chen},
  \bibinfo{person}{Omer Levy}, \bibinfo{person}{Mike Lewis},
  \bibinfo{person}{Luke Zettlemoyer}, {and} \bibinfo{person}{Veselin
  Stoyanov}.} \bibinfo{year}{2019}\natexlab{}.
\newblock \showarticletitle{Roberta: A robustly optimized bert pretraining
  approach}.
\newblock \bibinfo{journal}{\emph{arXiv preprint arXiv:1907.11692}}
  (\bibinfo{year}{2019}).
\newblock


\bibitem[\protect\citeauthoryear{Liu, Lin, Cao, Hu, Wei, Zhang, Lin, and
  Guo}{Liu et~al\mbox{.}}{2021}]%
        {cv3_2021}
\bibfield{author}{\bibinfo{person}{Ze Liu}, \bibinfo{person}{Yutong Lin},
  \bibinfo{person}{Yue Cao}, \bibinfo{person}{Han Hu}, \bibinfo{person}{Yixuan
  Wei}, \bibinfo{person}{Zheng Zhang}, \bibinfo{person}{Stephen Lin}, {and}
  \bibinfo{person}{Baining Guo}.} \bibinfo{year}{2021}\natexlab{}.
\newblock \showarticletitle{Swin transformer: Hierarchical vision transformer
  using shifted windows}. In \bibinfo{booktitle}{\emph{Proceedings of the
  IEEE/CVF international conference on computer vision}}.
  \bibinfo{pages}{10012--10022}.
\newblock


\bibitem[\protect\citeauthoryear{Nguyen, Lee, Rossi, Ahmed, Koh, and
  Kim}{Nguyen et~al\mbox{.}}{2018}]%
        {CTDNE2018}
\bibfield{author}{\bibinfo{person}{Giang~Hoang Nguyen},
  \bibinfo{person}{John~Boaz Lee}, \bibinfo{person}{Ryan~A Rossi},
  \bibinfo{person}{Nesreen~K Ahmed}, \bibinfo{person}{Eunyee Koh}, {and}
  \bibinfo{person}{Sungchul Kim}.} \bibinfo{year}{2018}\natexlab{}.
\newblock \showarticletitle{Continuous-time dynamic network embeddings}. In
  \bibinfo{booktitle}{\emph{Companion proceedings of the the web conference
  2018}}. \bibinfo{pages}{969--976}.
\newblock


\bibitem[\protect\citeauthoryear{Ramp{\'a}{\v{s}}ek, Galkin, Dwivedi, Luu,
  Wolf, and Beaini}{Ramp{\'a}{\v{s}}ek et~al\mbox{.}}{2022}]%
        {gtransformer_2022}
\bibfield{author}{\bibinfo{person}{Ladislav Ramp{\'a}{\v{s}}ek},
  \bibinfo{person}{Michael Galkin}, \bibinfo{person}{Vijay~Prakash Dwivedi},
  \bibinfo{person}{Anh~Tuan Luu}, \bibinfo{person}{Guy Wolf}, {and}
  \bibinfo{person}{Dominique Beaini}.} \bibinfo{year}{2022}\natexlab{}.
\newblock \showarticletitle{Recipe for a general, powerful, scalable graph
  transformer}.
\newblock \bibinfo{journal}{\emph{Advances in Neural Information Processing
  Systems}}  \bibinfo{volume}{35} (\bibinfo{year}{2022}),
  \bibinfo{pages}{14501--14515}.
\newblock


\bibitem[\protect\citeauthoryear{Rossi, Chamberlain, Frasca, Eynard, Monti, and
  Bronstein}{Rossi et~al\mbox{.}}{2020}]%
        {TGN2020}
\bibfield{author}{\bibinfo{person}{Emanuele Rossi}, \bibinfo{person}{Ben
  Chamberlain}, \bibinfo{person}{Fabrizio Frasca}, \bibinfo{person}{Davide
  Eynard}, \bibinfo{person}{Federico Monti}, {and} \bibinfo{person}{Michael
  Bronstein}.} \bibinfo{year}{2020}\natexlab{}.
\newblock \showarticletitle{Temporal graph networks for deep learning on
  dynamic graphs}.
\newblock \bibinfo{journal}{\emph{arXiv preprint arXiv:2006.10637}}
  (\bibinfo{year}{2020}).
\newblock


\bibitem[\protect\citeauthoryear{Sherstinsky}{Sherstinsky}{2020}]%
        {rnn2020}
\bibfield{author}{\bibinfo{person}{Alex Sherstinsky}.}
  \bibinfo{year}{2020}\natexlab{}.
\newblock \showarticletitle{Fundamentals of recurrent neural network (RNN) and
  long short-term memory (LSTM) network}.
\newblock \bibinfo{journal}{\emph{Physica D: Nonlinear Phenomena}}
  \bibinfo{volume}{404} (\bibinfo{year}{2020}), \bibinfo{pages}{132306}.
\newblock


\bibitem[\protect\citeauthoryear{Shi, Zheng, Ke, Shen, You, He, Luo, Liu, He,
  and Liu}{Shi et~al\mbox{.}}{2022}]%
        {gtransformer1_2022}
\bibfield{author}{\bibinfo{person}{Yu Shi}, \bibinfo{person}{Shuxin Zheng},
  \bibinfo{person}{Guolin Ke}, \bibinfo{person}{Yifei Shen},
  \bibinfo{person}{Jiacheng You}, \bibinfo{person}{Jiyan He},
  \bibinfo{person}{Shengjie Luo}, \bibinfo{person}{Chang Liu},
  \bibinfo{person}{Di He}, {and} \bibinfo{person}{Tie-Yan Liu}.}
  \bibinfo{year}{2022}\natexlab{}.
\newblock \showarticletitle{Benchmarking graphormer on large-scale molecular
  modeling datasets}.
\newblock \bibinfo{journal}{\emph{arXiv preprint arXiv:2203.04810}}
  (\bibinfo{year}{2022}).
\newblock


\bibitem[\protect\citeauthoryear{Souza, Mesquita, Kaski, and Garg}{Souza
  et~al\mbox{.}}{2022}]%
        {pint2022}
\bibfield{author}{\bibinfo{person}{Amauri Souza}, \bibinfo{person}{Diego
  Mesquita}, \bibinfo{person}{Samuel Kaski}, {and} \bibinfo{person}{Vikas
  Garg}.} \bibinfo{year}{2022}\natexlab{}.
\newblock \showarticletitle{Provably expressive temporal graph networks}.
\newblock \bibinfo{journal}{\emph{Advances in Neural Information Processing
  Systems}}  \bibinfo{volume}{35} (\bibinfo{year}{2022}),
  \bibinfo{pages}{32257--32269}.
\newblock


\bibitem[\protect\citeauthoryear{Trivedi, Farajtabar, Biswal, and Zha}{Trivedi
  et~al\mbox{.}}{2019}]%
        {DyRep2019}
\bibfield{author}{\bibinfo{person}{Rakshit Trivedi}, \bibinfo{person}{Mehrdad
  Farajtabar}, \bibinfo{person}{Prasenjeet Biswal}, {and}
  \bibinfo{person}{Hongyuan Zha}.} \bibinfo{year}{2019}\natexlab{}.
\newblock \showarticletitle{Dyrep: Learning representations over dynamic
  graphs}. In \bibinfo{booktitle}{\emph{International conference on learning
  representations}}.
\newblock


\bibitem[\protect\citeauthoryear{Van~Belle, Baesens, and De~Weerdt}{Van~Belle
  et~al\mbox{.}}{2023}]%
        {fraud2023}
\bibfield{author}{\bibinfo{person}{Rafa{\"e}l Van~Belle}, \bibinfo{person}{Bart
  Baesens}, {and} \bibinfo{person}{Jochen De~Weerdt}.}
  \bibinfo{year}{2023}\natexlab{}.
\newblock \showarticletitle{CATCHM: A novel network-based credit card fraud
  detection method using node representation learning}.
\newblock \bibinfo{journal}{\emph{Decision Support Systems}}
  \bibinfo{volume}{164} (\bibinfo{year}{2023}), \bibinfo{pages}{113866}.
\newblock


\bibitem[\protect\citeauthoryear{Vaswani, Shazeer, Parmar, Uszkoreit, Jones,
  Gomez, Kaiser, and Polosukhin}{Vaswani et~al\mbox{.}}{2017}]%
        {transformer2017}
\bibfield{author}{\bibinfo{person}{Ashish Vaswani}, \bibinfo{person}{Noam
  Shazeer}, \bibinfo{person}{Niki Parmar}, \bibinfo{person}{Jakob Uszkoreit},
  \bibinfo{person}{Llion Jones}, \bibinfo{person}{Aidan~N Gomez},
  \bibinfo{person}{{\L}ukasz Kaiser}, {and} \bibinfo{person}{Illia
  Polosukhin}.} \bibinfo{year}{2017}\natexlab{}.
\newblock \showarticletitle{Attention is all you need}.
\newblock \bibinfo{journal}{\emph{Advances in neural information processing
  systems}}  \bibinfo{volume}{30} (\bibinfo{year}{2017}).
\newblock


\bibitem[\protect\citeauthoryear{Veli{\v{c}}kovi{\'c}}{Veli{\v{c}}kovi{\'c}}{2023}]%
        {GNNs2023}
\bibfield{author}{\bibinfo{person}{Petar Veli{\v{c}}kovi{\'c}}.}
  \bibinfo{year}{2023}\natexlab{}.
\newblock \showarticletitle{Everything is connected: Graph neural networks}.
\newblock \bibinfo{journal}{\emph{Current Opinion in Structural Biology}}
  \bibinfo{volume}{79} (\bibinfo{year}{2023}), \bibinfo{pages}{102538}.
\newblock


\bibitem[\protect\citeauthoryear{Wang, Liu, Cheng, Liang, Gu, Li, Ding, Jiang,
  Shi, Qian, et~al\mbox{.}}{Wang et~al\mbox{.}}{2022}]%
        {graph_attention_layer2022}
\bibfield{author}{\bibinfo{person}{Hanrui Wang}, \bibinfo{person}{Pengyu Liu},
  \bibinfo{person}{Jinglei Cheng}, \bibinfo{person}{Zhiding Liang},
  \bibinfo{person}{Jiaqi Gu}, \bibinfo{person}{Zirui Li},
  \bibinfo{person}{Yongshan Ding}, \bibinfo{person}{Weiwen Jiang},
  \bibinfo{person}{Yiyu Shi}, \bibinfo{person}{Xuehai Qian}, {et~al\mbox{.}}}
  \bibinfo{year}{2022}\natexlab{}.
\newblock \showarticletitle{QuEst: Graph Transformer for Quantum Circuit
  Reliability Estimation}.
\newblock \bibinfo{journal}{\emph{arXiv preprint arXiv:2210.16724}}
  (\bibinfo{year}{2022}).
\newblock


\bibitem[\protect\citeauthoryear{Wang, Chang, Li, Chu, Li, Zhang, He, Song,
  Zhou, and Yang}{Wang et~al\mbox{.}}{2021b}]%
        {tcl2021}
\bibfield{author}{\bibinfo{person}{Lu Wang}, \bibinfo{person}{Xiaofu Chang},
  \bibinfo{person}{Shuang Li}, \bibinfo{person}{Yunfei Chu},
  \bibinfo{person}{Hui Li}, \bibinfo{person}{Wei Zhang},
  \bibinfo{person}{Xiaofeng He}, \bibinfo{person}{Le Song},
  \bibinfo{person}{Jingren Zhou}, {and} \bibinfo{person}{Hongxia Yang}.}
  \bibinfo{year}{2021}\natexlab{b}.
\newblock \showarticletitle{Tcl: Transformer-based dynamic graph modelling via
  contrastive learning}.
\newblock \bibinfo{journal}{\emph{arXiv preprint arXiv:2105.07944}}
  (\bibinfo{year}{2021}).
\newblock


\bibitem[\protect\citeauthoryear{Wang, Lyu, Li, Xia, Yang, Wang, Wang, Cui,
  Yang, Sun, et~al\mbox{.}}{Wang et~al\mbox{.}}{2021d}]%
        {apan2021}
\bibfield{author}{\bibinfo{person}{Xuhong Wang}, \bibinfo{person}{Ding Lyu},
  \bibinfo{person}{Mengjian Li}, \bibinfo{person}{Yang Xia},
  \bibinfo{person}{Qi Yang}, \bibinfo{person}{Xinwen Wang},
  \bibinfo{person}{Xinguang Wang}, \bibinfo{person}{Ping Cui},
  \bibinfo{person}{Yupu Yang}, \bibinfo{person}{Bowen Sun}, {et~al\mbox{.}}}
  \bibinfo{year}{2021}\natexlab{d}.
\newblock \showarticletitle{Apan: Asynchronous propagation attention network
  for real-time temporal graph embedding}. In
  \bibinfo{booktitle}{\emph{Proceedings of the 2021 international conference on
  management of data}}. \bibinfo{pages}{2628--2638}.
\newblock


\bibitem[\protect\citeauthoryear{Wang, Cai, Liang, Ding, Wang, Bhatia, and
  Hooi}{Wang et~al\mbox{.}}{2021a}]%
        {MATA2021}
\bibfield{author}{\bibinfo{person}{Yiwei Wang}, \bibinfo{person}{Yujun Cai},
  \bibinfo{person}{Yuxuan Liang}, \bibinfo{person}{Henghui Ding},
  \bibinfo{person}{Changhu Wang}, \bibinfo{person}{Siddharth Bhatia}, {and}
  \bibinfo{person}{Bryan Hooi}.} \bibinfo{year}{2021}\natexlab{a}.
\newblock \showarticletitle{Adaptive data augmentation on temporal graphs}.
\newblock \bibinfo{journal}{\emph{Advances in Neural Information Processing
  Systems}}  \bibinfo{volume}{34} (\bibinfo{year}{2021}),
  \bibinfo{pages}{1440--1452}.
\newblock


\bibitem[\protect\citeauthoryear{Wang, Chang, Liu, Leskovec, and Li}{Wang
  et~al\mbox{.}}{2021c}]%
        {caw2021}
\bibfield{author}{\bibinfo{person}{Yanbang Wang}, \bibinfo{person}{Yen-Yu
  Chang}, \bibinfo{person}{Yunyu Liu}, \bibinfo{person}{Jure Leskovec}, {and}
  \bibinfo{person}{Pan Li}.} \bibinfo{year}{2021}\natexlab{c}.
\newblock \showarticletitle{Inductive representation learning in temporal
  networks via causal anonymous walks}.
\newblock \bibinfo{journal}{\emph{arXiv preprint arXiv:2101.05974}}
  (\bibinfo{year}{2021}).
\newblock


\bibitem[\protect\citeauthoryear{Wu, Xu, Wang, and Long}{Wu
  et~al\mbox{.}}{2021}]%
        {ts2_2021}
\bibfield{author}{\bibinfo{person}{Haixu Wu}, \bibinfo{person}{Jiehui Xu},
  \bibinfo{person}{Jianmin Wang}, {and} \bibinfo{person}{Mingsheng Long}.}
  \bibinfo{year}{2021}\natexlab{}.
\newblock \showarticletitle{Autoformer: Decomposition transformers with
  auto-correlation for long-term series forecasting}.
\newblock \bibinfo{journal}{\emph{Advances in Neural Information Processing
  Systems}}  \bibinfo{volume}{34} (\bibinfo{year}{2021}),
  \bibinfo{pages}{22419--22430}.
\newblock


\bibitem[\protect\citeauthoryear{Xu, Ruan, Korpeoglu, Kumar, and Achan}{Xu
  et~al\mbox{.}}{2020}]%
        {TGAT2020}
\bibfield{author}{\bibinfo{person}{Da Xu}, \bibinfo{person}{Chuanwei Ruan},
  \bibinfo{person}{Evren Korpeoglu}, \bibinfo{person}{Sushant Kumar}, {and}
  \bibinfo{person}{Kannan Achan}.} \bibinfo{year}{2020}\natexlab{}.
\newblock \showarticletitle{Inductive representation learning on temporal
  graphs}.
\newblock \bibinfo{journal}{\emph{arXiv preprint arXiv:2002.07962}}
  (\bibinfo{year}{2020}).
\newblock


\bibitem[\protect\citeauthoryear{Xue, Zhong, Li, Chen, Zhai, and Kong}{Xue
  et~al\mbox{.}}{2022}]%
        {dynamicGraph2022}
\bibfield{author}{\bibinfo{person}{Guotong Xue}, \bibinfo{person}{Ming Zhong},
  \bibinfo{person}{Jianxin Li}, \bibinfo{person}{Jia Chen},
  \bibinfo{person}{Chengshuai Zhai}, {and} \bibinfo{person}{Ruochen Kong}.}
  \bibinfo{year}{2022}\natexlab{}.
\newblock \showarticletitle{Dynamic network embedding survey}.
\newblock \bibinfo{journal}{\emph{Neurocomputing}}  \bibinfo{volume}{472}
  (\bibinfo{year}{2022}), \bibinfo{pages}{212--223}.
\newblock


\bibitem[\protect\citeauthoryear{Zhang, Xiong, Li, Shan, Ren, and Zhu}{Zhang
  et~al\mbox{.}}{2021}]%
        {cope2021}
\bibfield{author}{\bibinfo{person}{Yao Zhang}, \bibinfo{person}{Yun Xiong},
  \bibinfo{person}{Dongsheng Li}, \bibinfo{person}{Caihua Shan},
  \bibinfo{person}{Kan Ren}, {and} \bibinfo{person}{Yangyong Zhu}.}
  \bibinfo{year}{2021}\natexlab{}.
\newblock \showarticletitle{CoPE: Modeling Continuous Propagation and Evolution
  on Interaction Graph}. In \bibinfo{booktitle}{\emph{Proceedings of the 30th
  ACM International Conference on Information \& Knowledge Management}}.
  \bibinfo{pages}{2627--2636}.
\newblock


\bibitem[\protect\citeauthoryear{Zhang, Xiong, Liao, Sun, Jin, Zheng, and
  Zhu}{Zhang et~al\mbox{.}}{2023}]%
        {tiger2023}
\bibfield{author}{\bibinfo{person}{Yao Zhang}, \bibinfo{person}{Yun Xiong},
  \bibinfo{person}{Yongxiang Liao}, \bibinfo{person}{Yiheng Sun},
  \bibinfo{person}{Yucheng Jin}, \bibinfo{person}{Xuehao Zheng}, {and}
  \bibinfo{person}{Yangyong Zhu}.} \bibinfo{year}{2023}\natexlab{}.
\newblock \showarticletitle{TIGER: Temporal Interaction Graph Embedding with
  Restarts} \emph{(\bibinfo{series}{WWW '23})}. \bibinfo{publisher}{Association
  for Computing Machinery}, \bibinfo{address}{New York, NY, USA},
  \bibinfo{pages}{478–488}.
\newblock
\showISBNx{9781450394161}
\urldef\tempurl%
\url{https://doi.org/10.1145/3543507.3583433}
\showDOI{\tempurl}


\bibitem[\protect\citeauthoryear{Zhou, Zhang, Peng, Zhang, Li, Xiong, and
  Zhang}{Zhou et~al\mbox{.}}{2021}]%
        {ts3_2021}
\bibfield{author}{\bibinfo{person}{Haoyi Zhou}, \bibinfo{person}{Shanghang
  Zhang}, \bibinfo{person}{Jieqi Peng}, \bibinfo{person}{Shuai Zhang},
  \bibinfo{person}{Jianxin Li}, \bibinfo{person}{Hui Xiong}, {and}
  \bibinfo{person}{Wancai Zhang}.} \bibinfo{year}{2021}\natexlab{}.
\newblock \showarticletitle{Informer: Beyond efficient transformer for long
  sequence time-series forecasting}. In \bibinfo{booktitle}{\emph{Proceedings
  of the AAAI conference on artificial intelligence}},
  Vol.~\bibinfo{volume}{35}. \bibinfo{pages}{11106--11115}.
\newblock


\bibitem[\protect\citeauthoryear{Zhou, Zheng, Nisa, Ioannidis, Song, and
  Karypis}{Zhou et~al\mbox{.}}{2022}]%
        {tgl2022}
\bibfield{author}{\bibinfo{person}{Hongkuan Zhou}, \bibinfo{person}{Da Zheng},
  \bibinfo{person}{Israt Nisa}, \bibinfo{person}{Vasileios Ioannidis},
  \bibinfo{person}{Xiang Song}, {and} \bibinfo{person}{George Karypis}.}
  \bibinfo{year}{2022}\natexlab{}.
\newblock \showarticletitle{Tgl: A general framework for temporal gnn training
  on billion-scale graphs}.
\newblock \bibinfo{journal}{\emph{arXiv preprint arXiv:2203.14883}}
  (\bibinfo{year}{2022}).
\newblock


\end{thebibliography}

\appendix

\end{document}